%% file: SparseDirectionalFilterDictionaries.tex
\documentclass[10pt,twocolumn,twoside]{article} 
\usepackage{simpleConference}
\usepackage[english]{babel}
\usepackage{times}
\usepackage{graphicx}
\graphicspath{{Images/}}
\usepackage{amsmath,amssymb}
\usepackage{caption}
\usepackage{subcaption}
\usepackage[hidelinks]{hyperref}
\usepackage{url}
\usepackage[utf8]{inputenc}
\usepackage{multirow}
\usepackage{booktabs}
\usepackage{color}
\usepackage{xcolor}
\usepackage{icomma} 
\usepackage{csquotes}
\usepackage{enumerate}
\usepackage{placeins}

\newtheorem{theorem}{Theorem}

\usepackage{fancyhdr}
\newcommand{\norm}[1]{\lVert#1\rVert}
\newcommand{\ip}[2]{\langle#1,#2\rangle}

\begin{document}

\title{Blind Image Inpainting with Sparse Directional Filter Dictionaries\\for Lightweight CNNs}

\author{\textbf{Jenny Schmalfuss} \\
University of Stuttgart\\
\and
\textbf{Erik Scheurer} \\
University of Stuttgart\\
\and
\textbf{Heng Zhao} \\
University of Houston\\
\and
\textbf{Nikolaos Karantzas} \\
University of Houston\\
\and
\textbf{Andrés Bruhn} \\
University of Stuttgart\\
\and
\textbf{Demetrio Labate} \\
University of Houston\\
\and
\{jenny.schmalfuss,andres.bruhn\}@vis.uni-stuttgart.de\\
erik.scheurer@simtech.uni-stuttgart.de\\
hzhao25@central.uh.edu\\
\{dlabate,nickos\}@math.uh.edu
}

\maketitle
\thispagestyle{empty}

\abstract{Blind inpainting algorithms based on deep learning architectures have shown a remarkable performance in recent years, typically outperforming model-based methods both in terms of image quality and run time. However, neural network strategies typically lack a theoretical explanation, which contrasts with the well-understood theory underlying model-based methods. In this work, we leverage the advantages of both approaches by integrating theoretically founded concepts from transform domain methods and sparse approximations into a CNN-based approach for blind image inpainting. To this end, we present a novel strategy to learn convolutional kernels that applies a specifically designed filter dictionary whose elements are linearly combined with trainable weights. Numerical experiments demonstrate the competitiveness of this approach. Our results show not only an improved inpainting quality compared to conventional CNNs but also significantly faster network convergence within a lightweight network design.\\}


\section{Introduction}

\begin{figure}[htb]
      \centering
      \def\svgwidth{\columnwidth}
      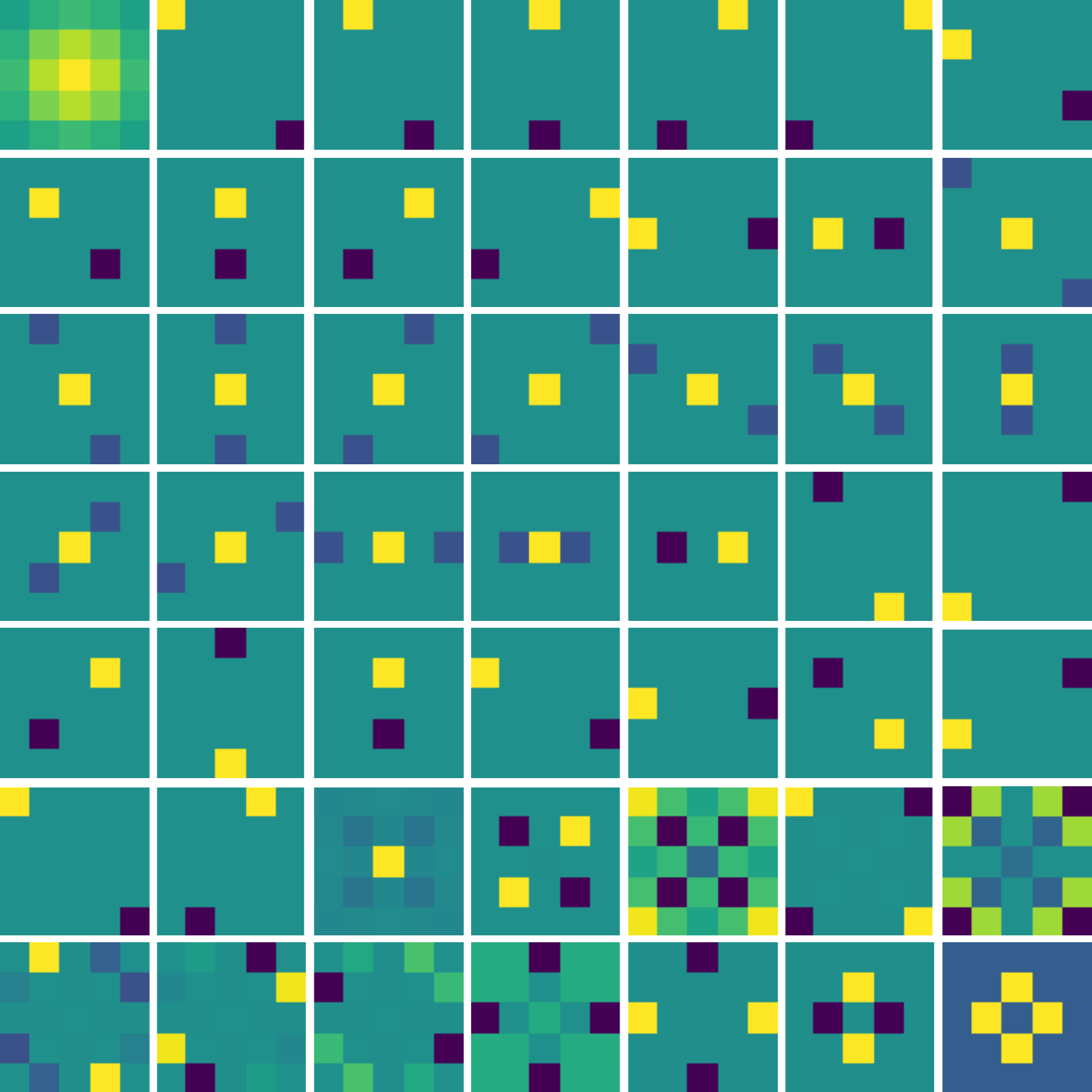
      \caption{The proposed $5\times 5$ Sparse Directional Parseval Frame (SDPF) dictionary. Filters from top left to bottom right: one low-pass filter; twelve first-order finite difference filters; twelve second-order finite difference filter; 24 filters for the Parseval frame completion.} \label{f.filters}
\end{figure}

Image inpainting is a longstanding problem in image processing, which aims to digitally remove visual corruptions from images that may be associated with scratches or other missing blocks of image information.
The inpainting problem can be divided into two formulations, \emph{blind} and \emph{non-blind} image inpainting, depending on the amount of a-priori knowledge about the image corruption.
For non-blind image inpainting, the location of the damage within the image is known and can be used within the algorithmic solution.
However, in this work, we focus on the more challenging \emph{blind inpainting} problem, which aims at recovering a missing region whose location is unknown.
Such location information can be missing when random pixels of an image are damaged, or when identifying the damage would otherwise require human interaction.
Due to the reduced amount of available information, the blind inpainting problem is {generally more difficult} to solve than non-blind inpainting.

Before the overwhelming success of neural networks for many image processing problems, the best performing strategies for blind image inpainting were model-based.
They heavily relied on a mathematical framework that was instrumental to solve the problem.
Due to this modeling aspect, classical image inpainting strategies (e.g., those based on variational or transform domain methods) are inherently predictable and explainable, which is often not true for learning-based approaches.
In this work, we develop a strategy to combine the high accuracy and fast evaluation times of Convolutional Neural Networks (CNNs) with the interpretability of model-based ideas.
Namely, we propose a novel notion of a receptive field layer that relies on the properties of a Parseval Frame dictionary (Fig.~\ref{f.filters}) specifically designed for image inpainting, which is combined with an appropriate sparsity constraint during network training.
As we argue below, this new strategy brings highly desirable properties from the theory of sparse image representations into the CNN model, making it not only more lightweight but also more explainable.

\subsection{Related work}

In the literature, the most successful inpainting strategies can be roughly grouped into three categories:
\begin{enumerate}[(i)]
\item\label{enum.reprmeth} \emph{Representation (or transform domain) methods} that formulate the inpainting problem as an optimization task in a transform domain~\cite{aharon_k-svd:_2006,cbs12,dong2012,elad2005,king_analysis_2014,shen2016};
\item \emph{PDE-based and Variational techniques} that recover missing data from the neighborhood points through a regularity criterion that might be associated with a PDE~\cite{esedoglu2002,shen2002,chan2005variational,chan2006,arias2011variational};
\item\label{enum.learnmeth} \emph{Learning-based strategies} such as convolutional neural networks (CNNs) that learn an end to end mapping from input images to inpainted images based on training data~\cite{xie_image_2012,cai_blind_2017,chaudhury_can_2017,deepblindimage,wang2020vcnet}.
\end{enumerate}
In the following, we discuss the state of the art for inpainting with (\ref{enum.reprmeth}) representation methods, (\ref{enum.learnmeth}) learning-based strategies as well as combinations of these seemingly alternative approaches, which is also the goal of this work.

Representation methods (\ref{enum.reprmeth}) model image inpainting as a signal restoration problem, where the image is represented as a superposition of a clean component and a \enquote{noisy} one.
It is then reasonable to assume that the clean component of the image has a sparse representation in some domain, e.g., in a wavelet space.
That is, it can be represented using relatively few representation coefficients.
Since the noise does not satisfy the same sparsity property, its energy is spread over the whole transform domain.
The clean image can then be recovered by identifying its sparse representation through $\ell_1$-norm minimization (in the transform domain).
This intuitive argument explains the critical role of \emph{sparsity} in image representation methods.

The study of efficient image representations has been the focus of an intense research starting with the introduction of wavelets in the late 1980's \cite{mallat1999} and continuing with the development of more advanced multiscale representations during the following two decades (cf.~the excellent review by Donoho et al.~\cite{DVDD1998} about the role of such representations in image processing).
Some of the most important developments in this area occurred with the introduction of curvelets~\cite{Candes2000} and shearlets~\cite{labate2005},
two multiscale methods that were shown to be provably sparser than traditional wavelets \cite{CanDon2004,Guo2007Optimal} for a large class of images called \emph{cartoon-like images}.
The hallmark of both methods is to combine the multiresolution structure of classical wavelets with superior directional sensitivity.
This directional sensitivity is achieved through the power of anisotropic scaling and the action of rotation or shear operators.

In parallel with the development of sparse representation methods, several sparsity-based algorithms for image inpainting were proposed in the literature; they include several methods based on wavelets \cite{cbs12,dong2012,shen2016} and shearlets \cite{king_analysis_2014} as well as methods such as K-SVD \cite{aharon_k-svd:_2006,mairal2007,mairal2008} that, rather than using a fixed dictionary as in the wavelet or shearlet case, build a dictionary adaptively from images.
While most of these results are focused on the algorithmic side, some research also investigated performance guarantees.
For instance, some results established a precise relationship between image inpainting of cartoon-like images and properties of the representation.
Due to their `geometric' properties, namely, their anisotropic support and directional sensitivity, shearlets were shown to offer a very convenient framework for inpainting  as they can provably fill larger gaps than wavelets in the class of cartoon-like images \cite{king_analysis_2014,guo2020}.

During the last five years or so, with the emergence of deep learning (\ref{enum.learnmeth}) in many areas of engineering and applied mathematics, deep learning methods have gained increasing recognition also in image inpainting due to their very competitive performance.
Such methods have been especially effective to address non-blind inpainting, with earlier works using simple architectures like multilayer perceptrons~\cite{jiang_mask-specific_2014} or encoder-decoder structures~\cite{pathak_context_2016}.
Later research focused on developing alternatives for convolutions that specifically use the corruption's location, such as partial convolutions~\cite{liu_image_2018} or gated convolutions~\cite{Yu_2019_ICCV}.
Most recent approaches for non-blind image inpainting use generative adversarial networks (GANs) in order to inpainting missing regions in ways that are visually hard to discriminate from similar images~\cite{Yu_2018_CVPR,Yu_2019_ICCV,yi2020contextual}.
The main limitation of such methods is that their performance if highly dependent on the type of images used for training.
In addition, they usually fail if the image location to be inpainted if unknown.
By contrast, a much smaller number of methods were proposed to address the more challenging blind inpainting problem.
Existing methods often use encoder-decoder structures based on CNN architectures~\cite{cai_blind_2017,chaudhury_can_2017} that may contain residual blocks to improve performance~\cite{deepblindimage}.
The most advanced and best performing schemes in the literature for blind inpainting adopt a two-stage approach where the first stage of the algorithm estimates the location of image corruptions and the second stage applies a non-blind inpainting pass on the detected regions~\cite{wang2020vcnet}.
The improved performance comes at the cost of a significantly higher network complexity.

While deep neural networks have demonstrated impressive results and often outperform conventional methods, one major concern is the lack of interpretability due to their black-box nature~\cite{interpretability}.
Therefore, there is an increasing effort to include principles from model-based methods into deep learning for an improved interpretability.
For instance, Xie et al.~\cite{xie_image_2012} proposed a network-based image restoration approach consisting of stacked denoising autoencoder (SDA) that takes inspirations from the K-SVD algorithm~\cite{aharon_k-svd:_2006}.
Inspired by the structure of a sparse representation method, each denoising autoencoder block is a two-layer neural network trained to reconstruct a clean image from a corrupted one with the inner layer representation constrained to be sparse.
Similarly, Chaudhury and Roy~\cite{chaudhury_can_2017} proposed a CNN for image restoration (IRCNN) where the hidden layers are designed to learn a data-driven sparse representation.
Another method to bring principles of classical image representation into neural network designs are the Structured Receptive Field Networks (SRFN), where convolutional filters are built as linear combinations from a pre-defined dictionary, and only the coefficients of the representation are learned during training~\cite{Jacobsen2016}.
While this idea was originally proposed to learn expressive feature representations in scenarios with limited training data, we adapt and refine this idea to build our approach for image inpainting.

\subsection{Contribution}

In this work, we introduce a novel blind inpainting strategy that leverages the computational efficiency of a CNN along with the interpretability of mathematical representation methods.
For that, we adopt the SRFN idea where each convolutional filter is a linear combination of elements from a fixed dictionary, where the coefficients of the linear combination are learned during training.
Within this framework, we take advantage of the  successful theory of multiscale directional representations to build a new discrete dictionary that is especially effective for image inpainting.
Below are the main contributions of our approach.

\begin{enumerate}[(1)]
\item We design a new dictionary of filters to provide efficient representations for salient features such as edges and corners in natural images. Our filter design is based on a recently proposed mathematical framework for the construction of Parseval frames with compact support~\cite{atreas_design_2018}.
\item We include a sparsity constraint during the training that is inspired by the sparsity norms used in model-based representation methods.
\item We implement our inpainting strategy using a simple transform CNN architecture \cite{chaudhury_can_2017}. After examining the most effective placement of receptive field layers, we select two lightweight architectures.
\item We run numerical experiments to demonstrate the capabilities of our method as compared to
state-of-the-art methods for blind inpainting. Specifically,
      \begin{enumerate}
      \item we demonstrate the learning capabilities of our receptive field layers and provide an interpretation of  their capabilities in terms of image representation;
      \item we show that our filter learning strategy provides more than merely a good layer initialization and demonstrate the efficacy of our approach throughout the complete training process;
      \item we experimentally confirm that our network strategy significantly reduces the amount of training data required for high-quality inpainting results.
      \end{enumerate}
\end{enumerate}
We remark that another application of the SRFN idea in the context of hyperspectral classification was presented by some of the authors in~\cite{spie2019}, but with a very different rationale, network design and algorithm.

\section{Method}

We formulate image inpainting as an inverse problem that aims to recover an image~$x$ from its corrupted version $y=x+w$. A solution of this problem is found by solving
\begin{equation}  \label{eq.den}
\hat x = \arg \min_x \norm{y - x }_2^2 +\lambda \Phi(x),
\end{equation}
where $\lambda$ is a trade-off parameter and $\Phi$ is a regularization operator enforcing some condition on the solution, e.g., sparsity.

Rather than  solving Equation~\eqref{eq.den} directly through image representation methods and optimization techniques, here we opt for a learning-based approach.
This choice is motivated by the efficiency of learning-based methods to address a specific corruption process, e.g., noise removal, and their fast evaluation speed compared to model-based approaches.
Additionally, we include concepts from representation methods to endow our CNN approach with interpretability.
In contrast to existing methods that try to combine the advantages of learning- and model-based methods by mimicking the structure representation methods at the architectural level \cite{xie_image_2012,chaudhury_can_2017}, our approach acts at the layer level.

\subsection{Method overview}

\begin{figure*}[t]
    \centering
    \small
    \def\svgwidth{.7\textwidth}
    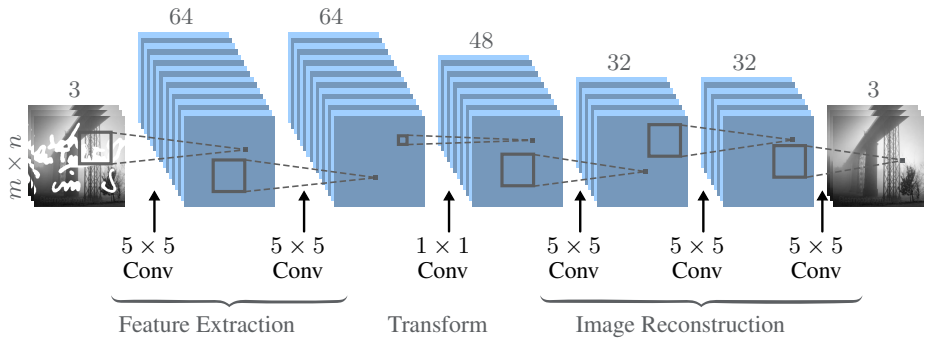%
    \caption{Baseline network architecture, adapted from~\cite{chaudhury_can_2017}. %
      Similar to classical transform domain methods, this architecture includes 3 blocks: feature extraction, transform and reconstruction. Our geometric biased, sparse filters from Fig.~\ref{f.filters} have the largest impact if used in the first two layers.}%
    \label{f.IRCNN_architecture}%
\end{figure*}

The architectural foundation for our network design is a slim, fully convolutional network architecture similar to the Image Restoration CNN (IRCNN) by Chaudhury and Roy~\cite{chaudhury_can_2017}.
It consists of a sequence of convolutional layers with ReLU activations, and can therefore process images of any input dimensions~\cite{long_fully_2015}.
As shown in Fig.~\ref{f.IRCNN_architecture}, the network architecture forms three sections that resemble the process of a classical representation method: feature extraction, nonlinear dimensionality reduction and clean image reconstruction.
We select this network architecture due to this structural resemblance to representation methods as well as its simplicity.
However we remark that our following strategy of expressing a convolutional filter
as linear combination from a pre-designed sparse dictionary can be applied to any existing CNN architecture.

Following the SRFN idea, we assume that any $5 \times 5$ convolutional filter $\{S_k\}$ of our network is expressed as linear combination of $5 \times 5$ basis filters $B_i$, $i=1, \dots L$ that are taken from an appropriate dictionary. Hence, any convolutional filter  $S_k$ is of the form
\begin{equation}\label{equ.SRFN}
S_k = \sum_{i=1}^L \alpha_{k,i} B_i,
\end{equation}
where the filter coefficients $\alpha_{k,i}$ are learned during training. With respect to the original SRFN method~\cite{Jacobsen2016}, we introduce two innovative features that, combined with our selection of network architecture, are designed to reflect some fundamental idea from the sparsity-based approach to inpainting.

One major novelty of our approach is to build our filter dictionary as a tight frame consisting of directional filters based on the theory of shearlets.
Guided by the theoretical insight that shearlet-based inpainting algorithms achieved state-of-the-art performance~\cite{king_analysis_2014}, our strategy is to select the convolutional filters of our CNN from a shearlet-like dictionary consisting of filters with high directional sensitivity.
We remark that we cannot use the original shearlet filters that are defined in the Fourier domain \cite{Guo2007Optimal,king_analysis_2014} nor its space-domain variants \cite{Kutyniok2010Compact} due to their large support.
To implement them in a CNN, we need a filter with small support.
Therefore, we designed a shearlet-like filter dictionary consisting of $5 \times 5$ matrices with the following properties:
\begin{enumerate}[(i)]
\item it forms a Parseval frame (completeness),
\item it produces discrete directional differentiation in all directions (edge detection),
\item the filters have few non-zero entries (fast computation).
\end{enumerate}

In this paper, we solve the filter design problem using the theory of \emph{compactly supported directional framelets}~\cite{atreas_design_2018}, recently proposed by one of the authors.
As shown in Fig.~\ref{f.filters}, our dictionary contains  $5\times5$ filters with a pronounced directional response that are highly efficient to capture edges and sharp transitions in images.
In the next section, we illustrate the dictionary construction in detail.

Another novelty of our approach is to impose a sparsity constraint during training that limits the number of dictionary elements allowed in any linear combination of filters.
That is, in Equation \eqref{equ.SRFN}, we only allow the sum to contain a small number of terms, e.g., three.
This condition can be interpreted as a \emph{geometric constraint}.
The elements of our filter dictionary include low-pass filters and edge detectors along various discrete orientations.
Therefore, a linear combination of a few filters of this type generates a kernel acting as a low pass filter or an edge detector along selected orientations.
As a consequence, while the weights of the network layer are still determined by data, our filter construction strategy results in convolutional kernels that are interpretable: They may act as direction-selective edge detectors or averaging operators.
We also remark that, due to the sparsity constraint, this approach requires significantly fewer trainable weights than a standard convolution, where a weight for every filter pixel has to be learned.

In the following, we call the dictionary built for our network a \emph{Sparse Directional Parseval Frame (SDPF) dictionary} and a convolutional layer build from this SDPF dictionary using Equation~\eqref{equ.SRFN} (possibly with the sparsity constraint) a \emph{SDPF constrained receptive field layer}.

\subsection{Filter design}

Our filter design approach adapts a method recently proposed by one of the authors in \cite{atreas_design_2018}, enabling the construction of discrete frames with prescribed support size that are easy to implement numerically.
In brief, this idea consists of choosing a discrete low-pass filter and a set of high-pass filters.
Additional high-pass filters are added until the combined set forms a frame or a Parseval frame.
There is some flexibility in choosing the high pass filters and this can be exploited to endow the frame with desirable properties.

For the formal definitions, recall that a collection $\{v_i\}$ in a Hilbert space $\mathcal{H}$ is a \emph{frame} if there are \emph{lower} and \emph{upper frame bounds} $a,b$ with  $0 < a \le b < \infty$ such that
\begin{equation}
a \, \norm{v}^2 \le \sum_i \lvert \ip{v}{v_i} \rvert^2 \le b \, \norm{v}^2
\end{equation}
for all elements $v \in \mathcal{H}$.
A frame is a \emph{Parseval frame} if $a=b=1$.
The Parseval frame condition generalizes the notion of orthonormal basis and ensures that any element $v \in \mathcal{H}$ can be expressed as a linear combination $v= \sum_n {\alpha_i(v)} \, v_i$ of frame elements.
In other words, the system $\{v_i\}$ is complete in $\mathcal{H}$.

The first step in the construction of our $5 \times 5$ Parseval frame is the selection of a low pass filter with positive coefficients.
We choose the 1-dimensional low pass filter associated with the fourth order cardinal $B$-spline,
\begin{equation}
\mu_{0}(\gamma)=\left(\frac{1+e^{2\pi i\gamma}}{2}\right)^{4}, \quad \gamma = 0, \dots 4,
\end{equation}
so that, by taking its tensor product, the corresponding $5 \times 5$ low-pass filter is
\begin{equation}
h_{0}=\frac{1}{64}\left(\begin{smallmatrix}
1&4&6&4&1\\4&16&24&16&4\\6&24&36&24&6\\4&16&24&16&4\\1&4&6&4&1
\end{smallmatrix}\right).
\end{equation}

We recall the following theorem from \cite{atreas_design_2018}:
\begin{theorem}   \label{conditionhigh}
      Let $h_0$ be a tensor product of univariate B-splines of order $n$ and $a$ be the vectorization of $h_{0}$.
      Let $H_{0}(\xi)=c W(\xi)$, where
      \begin{equation}
      c=(\sqrt{a_{0}},\sqrt{a_{1}},\ldots,\sqrt{a_{N-1}}),
      \end{equation}
      $N = (n+1)\times (n+1)$ and
      \begin{equation}
      W(\xi)=\left(\sqrt{a_{0}},\sqrt{a_{1}}e^{2\pi i \xi},\ldots,\sqrt{a_{N-1}}e^{2\pi i (N-1)\xi}\right)^\top .
      \end{equation}
      If $v \geq \max\{N, 2^n-1\}$ and $Y$ is a $v\times N$ real-valued matrix such that the rows of $\left( 
      \begin{tabular}{c} c \\
      Y
      \end{tabular} \right)$ form a Parseval frame in $\mathbb{R}^N$ and all rows of $Y$ are perpendicular to  $c$, and $h_{1,i}(\xi)=Y_i W(\xi)$, where $Y_i$ is the i-th row of $Y$, then the rows of the $v \times N$ matrix $B = Y diag(c)$ hold the high pass filter coefficients inducing a Parseval frame of  $L^{2}(\mathbb{R}^{N})$.
\end{theorem}

We will apply Theorem~\ref{conditionhigh} with $n=4$ to build a Parseval frame with $5\times 5$ filter elements in $\mathbb{R}^{25}$ ($N =25$).
According to Theorem~\ref{conditionhigh}, in $Y$ each high-pass filter occupies a single row and we can hand pick any filters that we wish, provided that each row of $Y$ is perpendicular to $c$.
This flexibility allows us to choose desirable properties such as directional sensitivity at specific orientations, which is motivated by the properties of the shearlet representation \cite{guo2020}.
Hence, in our construction, we choose a set of first-order central difference filters $h_{i}$, $i=1,\ldots,12,$ oriented at all possible discrete orientations on the $5\times5$ grid
\begin{align*}
h_{1}=&\left(\begin{smallmatrix}0&0&0&0&1\\
0&0&0&0&0\\
0&0&0&0&0\\
0&0&0&0&0\\
-1&0&0&0&0\\
\end{smallmatrix}\right),&
h_{2}=&\left(\begin{smallmatrix}0&0&0&1&0\\
0&0&0&0&0\\
0&0&0&0&0\\
0&0&0&0&0\\
0&-1&0&0&0\\
\end{smallmatrix}\right),\\
&\cdots &&\cdots\\
h_{11}=&\left(\begin{smallmatrix}0&0&0&0&0\\
0&0&0&0&0\\
-1&0&0&0&1\\
0&0&0&0&0\\
0&0&0&0&0\\
\end{smallmatrix}\right),&
h_{12}=&\left(\begin{smallmatrix}0&0&0&0&0\\
0&0&0&0&0\\
0&-1&0&1&0\\
0&0&0&0&0\\
0&0&0&0&0\\
\end{smallmatrix}\right).
\end{align*}
Additionally, we select a set of second-order central difference filters $h_{i}$, $i=13,\ldots,24,$
\begin{align*}
h_{13}=&\left(\begin{smallmatrix}0&0&0&0&-1\\
0&0&0&0&0\\
0&0&2&0&0\\
0&0&0&0&0\\
-1&0&0&0&0\\
\end{smallmatrix}\right),&
h_{14}=&\left(\begin{smallmatrix}0&0&0&-1&0\\
0&0&0&0&0\\
0&0&2&0&0\\
0&0&0&0&0\\
0&-1&0&0&0\\
\end{smallmatrix}\right),\\
&\cdots &&\cdots\\
h_{23}=&\left(\begin{smallmatrix}0&0&0&0&0\\
0&0&0&0&0\\
-1&0&2&0&-1\\
0&0&0&0&0\\
0&0&0&0&0\\
\end{smallmatrix}\right),&
h_{24}=&\left(\begin{smallmatrix}0&0&0&0&0\\
0&0&0&0&0\\
0&-1&2&-1&0\\
0&0&0&0&0\\
0&0&0&0&0\\
\end{smallmatrix}\right).
\end{align*}

From here, the idea is to build a matrix
$Y = \begin{pmatrix}
      D_{1}(\lambda^{*}) \\
      D_2
\end{pmatrix}
$
where $D_{1}(\lambda^{*})$ includes our chosen first- and second order filters (up to rescaling) and $D_2$ is a suitable completion matrix ensuring that  $B = Y \mathop{diag}(c)$ holds the high pass filter coefficients of the Parseval frame.
Such a completion matrix $D_2$ can be found if the singular values $s_\nu$ of
$Q = \begin{pmatrix}
      c \\
      D_{1}(\lambda^{*})
\end{pmatrix}
$
satisfy $s_\nu \leq 1$ (Lemma~3.1 in \cite{atreas_design_2018}).

To this end, we first need to define $D_1(\lambda)$ and then find $\lambda = \lambda^*$ such that $Q$ satisfies $s_\nu \leq 1$.
We convert the matrix filters $h_i$ ($i=1,\hdots,25$) into vectors by defining the map $\Lambda : \mathbb{R}^{5 \times 5} \rightarrow \mathbb{R}^{25}$  where
\begin{equation}
\Lambda \left( h_i \right) = \left(h_i^{5,1},\hdots,h_i^{5,5},\hdots, h_i^{1,1},\hdots,h_i^{1,5} \right)
\end{equation}
and, for a real variable $\lambda$  we let
\begin{equation}
d\left(\lambda, h_i \right):=\lambda \left( \frac{\Lambda \left(h_i\right)_{k}}{c_{k}}\right)_{k=1}^{25}.
\end{equation}
Given $c = (c_{k})_{k=1}^{25}$, and assuming an element-wise division in the definition of $d(\lambda, h_i)$, we define the matrix $D_{1}(\lambda) \in {\mathbb R}^{24 \times 25}$ as the matrix whose rows are the $d(\lambda, h_i)$ vectors associated with the 24 high-pass filters $h_i$ given above.

To find the value $\lambda = \lambda^{*}$ such that the singular values $s_\nu$ of
$Q = \begin{pmatrix}
      c\\
      D_{1}(\lambda^{*})
\end{pmatrix}
$
satisfy $s_\nu \le 1$, we solve
\begin{align}
\max \quad
& \text{trace} (c^\top c+D_{1}(\lambda)^\top D_{1}(\lambda))\\
\text{s.t.}\quad
& \left \| c^\top c + D_{1}(\lambda)^\top D_{1}(\lambda) \right \| \leq 1 .
\end{align}
As noted before, provided that the singular values of $Q \in {\mathbb R}^{25 \times 25}$ satisfy $s_\nu \le 1$, one can show (Lemma~3.1 in~\cite{atreas_design_2018}) that we  find a completion matrix $D_{2}$ for which rows of
\begin{equation}
\begin{pmatrix}
c\\
Y
\end{pmatrix} =
\begin{pmatrix}
c\\
D_{1}(\lambda^{*})\\
D_{2}
\end{pmatrix} \in \mathbb{R}^{(\upsilon+1)\times 25}, \quad \upsilon \geq 24
\end{equation}
form a Parseval frame for $\mathbb{R}^{25}$.

Thus the existence of a matrix completion $D_2$ to obtain the Parseval frame for $\mathbb{R}^{25}$ is ensured.
Conveniently the proof of Lemma~3.1 in~\cite{atreas_design_2018} is constructive and therefore provides clear instructions on how to construct $D_2$:
First, we perform a Singular Value Decomposition (SVD) on $Q$, which gives $Q = U\Sigma_{1}V^\top$ with $U \in \mathbb{R}^{25 \times 25}$, $V\in \mathbb{R}^{25 \times 25}$
and
\begin{equation}
\Sigma_{1} = \text{diag}(\sigma_{1},\hdots, \sigma_{25}) \in \mathbb{R}^{25 \times 25}.
\end{equation}
Given the singular values $\sigma_{1},\hdots, \sigma_{25}$ and matrix $V$ from $Q$, the completion matrix $D_{2}$ is constructed as $D_{2}= \Sigma_{2}V^\top \in \mathbb{R}^{25 \times 25}$ with
\begin{equation}
\Sigma_{2} = \text{diag}(\sqrt{1-\sigma_{1}^{2}},\hdots,\sqrt{1-\sigma_{25}^{2}}) .
\end{equation}
 From the result of the SVD, we have $\sigma_{1} = 1$, then $\sqrt{1-\sigma_{1}^{2}} = 0$ and the matrix $\Sigma_{2} \in \mathbb{R}^{25\times 25}$ reduces to
\begin{equation}
\Sigma_{2} = \text{diag}(0,\sqrt{1-\sigma_{2}^{2}},\hdots,\sqrt{1-\sigma_{25}^{2}})
\end{equation}
so that the first row of $D_{2}$ will be a zero vector.
Therefore, we get 24 new filters to complete the Parseval frame for $\mathbb{R}^{25}$, and we obtain a high-pass filter matrix as
\begin{equation}
B = \begin{pmatrix}
D_{1}(\lambda^{*}) \\
D_2
\end{pmatrix} \text{diag}(c).
\end{equation}
Figure~\ref{f.filters} shows that our $5 \times 5$ filter dictionary includes a low-pass filter, 24 first- and second-order finite difference filters, followed by 24 additional filters required to obtain a Parseval frame.

We remark that we could use a similar idea to build filters of different support size, e.g., $3 \times 3$ or $7 \times 7$.
In this work, we selected the size $5 \times 5$ since it offers a good compromise between efficiency and complexity.
Filters with shorter support (e.g., $3 \times 3$) would have reduced geometric sensitivity (e.g., we can only handle 4 orientations in the $3 \times 3$ grid).
Filter with larger support would be more complex due to more orientations and more dictionary elements.

\section{Results}\label{sec:Results}

The experimental evaluation of our image inpainting algorithm is divided into four parts.
In the first part, we discuss how to select the best configuration of SPDF constrained receptive field layers in our network for image inpainting.
The second part provides a deeper analysis of the learned filters, and the impact of SPDF constrained receptive field layers on the training process.
In the third part, we analyze the amount of training data that is required to effectively train a network with SPDF constrained receptive field layers as compared to a conventional CNN.
Lastly we measure the inpainting quality and run time of our top performing network configurations in comparison to state-of-the-art inpainting methods.
Preceding the experimental analysis we describe the imaging dataset used for our experiments.

\subsection{Experimental data set}

The dataset used for our experiments consists of $225,100$ images with  $256 \times 256$ pixels from the  \emph{Places} data set~\cite{zhou_places:_2018}.
We overlaid the images with handwriting masks (line width about~10 pixels) extracted from scanned pages, cf.~Fig.~\ref{fig.comparison} for examples.
Note that we obtained our mask library by rotating a set of $56,275$ handwriting masks by $0^\circ$, $90^\circ$, $180^\circ$ and $270^\circ$, resulting in a total of $225,100$ masks.

Our handwriting masks provide a variable coverage of a given image, ranging from $1$\% to $25$\% of the total number of pixels in an image.
Table~\ref{tab:ImCoverage} lists the distribution of masks by the size of the area they occlude, and how they are split into training and test sets.
Note that not enough samples were available in the coverage range $20$-$25$\%, to allow an even distribution of test samples.
\begin{table}[thb]
      \center
  \caption{Number of handwriting images in the training and test sets, grouped by occlusion area. }
  \label{tab:ImCoverage}
      \scalebox{1}{
      \begin{tabular}{@{\ }lrrr@{\ }}
          \toprule
          \bf Occlusion & \bf Training & \bf Test\\
          \midrule
          $0$-$5$\%  & $100,000$ & $1,000$ \\
          $5$-$10$\%  & $100,000$ & $1,000$ \\
          $10$-$15$\%  & $10,000$ & $1,000$ \\
          $15$-$20$\%  & $10,000$ & $1,000$ \\
          $20$-$25$\%  & $1,000$ & $100$ \\
      \bottomrule
      \end{tabular}}
\end{table}

\subsection{Network configuration and training}\label{subsec:Network configuratio and training}

\begin{table*}[thb]
   \center
\caption{Inpainting comparison for network configurations that use exactly one SDPF constrained receptive field layer. Performance values are computed on the test set, after the network was trained on $5,000$ training images for $100$ epochs; results are averaged over $10$ training runs.}
\label{tab: setupcomparison 5000}\small
   \scalebox{1.}{
\begin{tabular}{@{\ }lrcccc@{\ }}
\toprule
\multirow{ 2}{*}{\textbf{Configuration}} &  \multirow{ 2}{*}{\textbf{Param}.} &  \textbf{MSE}               &  \textbf{L$_1$}             &  \multirow{ 2}{*}{\textbf{PSNR}} &  \multirow{ 2}{*}{\textbf{SSIM}}  \\
              &         &  ($\times 10^{-2}$) &  ($\times 10^{-2}$) &       &   \\
\midrule
\texttt{C-C-c-C-C-C} (IRCNN) & 172,113         &  0.1437          &          1.3949 &       30.5622    &          0.9437 \\
\texttt{B-C-c-C-C-C} (GBCNN) & 170,705         &  \textbf{0.1074} & \textbf{0.9458} & \textbf{32.2214} & \textbf{0.9552} \\
\texttt{C-B-c-C-C-C}         & \textbf{82,001} &  0.1180          &          1.0838 &       31.7225    &          0.9520 \\
\texttt{C-C-c-B-C-C}         & 138,321         &  0.1283          &          1.2192 &       31.1119    &          0.9469 \\
\texttt{C-C-c-C-B-C}         & 149,585         &  0.1359          &          1.2727 &       30.8522    &          0.9430 \\
\texttt{C-C-c-C-C-B}         & 171,409         &  0.1366          &          1.3159 &       30.7806    &          0.9442 \\
\bottomrule
\end{tabular}%
}
\end{table*}

To identify the best network configuration for inpainting, we systematically tested different architectures where any convolutional layer is either a SDPF constrained receptive field layer or a conventional convolutional layer.
In the first case, any convolutional kernel is a linear combination of only 3 filters selected from the SPDF dictionary, that is, we imposed a sparsity constraint with sparsity 3.
In our implementation, the sparsity level is treated as a hyperparameter of the neural network, and 3 filters are randomly chosen per sparse filter when the network is initialized.
We tested several filter numbers and heuristically found 3 filters to work best for our dataset.

We trained the networks on a reduced training set consisting of~$5,000$ images ($1,000$ per occlusion area range) to limit the computational budget.
The CNNs were implemented using Tensorflow~\cite{abadi_tensorflow:_2016} with the Adam optimizer~\cite{kingma_adam:_2014} (using standard settings) and trained for~$100$ epochs with batch size~$10$.

To describe the network architectures, in the following we denote a conventional $5\times 5$ convolutional layer by~\texttt{C}, while we denote by~\texttt{c} a conventional $1\times 1$ convolutional layer.
By contrast, we denote by~\texttt{B} a SDPF constrained receptive field layer with sparsity 3.
As our dictionary is constructed from $5\times 5$ filters, the third layer \texttt{c} is never replaced by a receptive field layer.
Note that the configuration with only convolutional layers (\texttt{C-C-c-C-C-C}) is the IRCNN from~\cite{chaudhury_can_2017}.

Table~\ref{tab: setupcomparison 5000} reports the image inpainting performance for multiple network configurations that are derived from the CNN architecture in Fig.~\ref{f.IRCNN_architecture} by using either conventional convolutional layers or SDPF constrained receptive field layers.
We note that all configurations with a receptive field layer~(\texttt{B}) outperform the original IRCNN architecture for all measured metrics, indicating the improvement due to our modified design and despite the reduced number of trainable parameters.

{Among the various configurations, the best inpainting quality is achieved with the first layer being a SDPF constrained receptive field layer and the remaining layers being conventional convolutional layers (\texttt{B-C-c-C-C-C}).
For brevity, we call this configuration \textit{Geometric-Biased CNN} (GBCNN), due to the geometric bias (high directional sensitivity) that is imposed by the combination of SDPF filters and sparsity constrained in the first layer.
We explain the excellent performance of this configuration with the improved ability of the SDPF constrained receptive field layer to respond to geometric cues in images, such as edges or corners - an ability which is particularly effective in the first layer(s).}
This is supported the performance measures in Table~\ref{tab: setupcomparison 5000}, which show a decreasing inpainting performance with the depth of the SDPF filters in the network.
Out of all network configurations that use SDPF filters, \texttt{C-C-c-C-C-B} yields the worst performance, even though it still outperforms the fully convolutional IRCNN architecture.
Also, the weakening effect of the SDPF filters with increasing network depth appears to be independent of the network's parameter count (and hence its expressiveness), indicating that the improved performance indeed comes from the introduction of model assumptions into the network.

\begin{table*}[thb]
\center
\caption{Inpainting comparison of network configurations based on the GBCNN architecture, trained on $5,000$ images for $100$ epochs with results averaged over $10$ training runs. Best results displayed in bold.}
\label{tab.setupcomparison_firstlaymix}\small
\scalebox{1.}{
\begin{tabular}{@{\ }lrcccc@{\ }}
\toprule
\multirow{ 2}{*}{\textbf{Configuration}} &  \multirow{ 2}{*}{\textbf{Param}.} &  \textbf{MSE}               &  \textbf{L$_1$}             &  \multirow{ 2}{*}{\textbf{PSNR}} &  \multirow{ 2}{*}{\textbf{SSIM}}  \\
              &         &  ($\times 10^{-2}$) &  ($\times 10^{-2}$) &       &   \\
\midrule
\texttt{B-C-c-C-C-C} (GBCNN) & 170,705         &    \textbf{0.1074} &    \textbf{0.9459} &   \textbf{32.2213} &     \textbf{0.9552}\\
\midrule
\texttt{B-C-c-C-C-B} &                    170,001 &            0.1113  &    \textbf{0.9867} &   \textbf{32.0204} &             0.9532 \\
\texttt{B-C-c-C-B-C} &                    148,177 &            0.1138  &            1.0383  &           31.8524  &             0.9519 \\
\texttt{B-C-c-C-B-B} &                    147,473 &            0.1230  &            1.0890  &           31.5003  &             0.9487 \\
\texttt{B-C-c-B-C-C} &                    136,913 &            0.1160  &            1.0469  &           31.7855  &             0.9519 \\
\texttt{B-C-c-B-C-B} &                    136,209 &            0.1183  &            1.0465  &           31.6908  &             0.9505 \\
\texttt{B-C-c-B-B-C} &                    114,385 &            0.1255  &            1.1303  &           31.3345  &             0.9479 \\
\texttt{B-C-c-B-B-B} &                    113,681 &            0.1441  &            1.2946  &           30.5393  &             0.9405 \\
\midrule
\texttt{B-B-c-C-C-C} (GBCNN\nobreakdash-L) &           80,593 &    \textbf{0.1109} &            1.0113  &           31.9925  &     \textbf{0.9538}\\
\texttt{B-B-c-C-C-B} &                     79,889 &            0.1166  &            1.0451  &           31.7482  &             0.9513 \\
\texttt{B-B-c-C-B-C} &                     58,065 &            0.1391  &            1.3125  &           30.7628  &             0.9430 \\
\texttt{B-B-c-C-B-B} &                     57,361 &            0.1294  &            1.1457  &           31.1907  &             0.9462 \\
\texttt{B-B-c-B-C-C} &                     46,801 &            0.1184  &            1.0562  &           31.6847  &             0.9508 \\
\texttt{B-B-c-B-C-B} &                     46,097 &            0.1244  &            1.0977  &           31.4052  &             0.9484 \\
\texttt{B-B-c-B-B-C} &                     24,273 &            0.1292  &            1.1401  &           31.1820  &             0.9462 \\
\texttt{B-B-c-B-B-B} &            \textbf{23,569} &            0.1484  &            1.3634  &           30.3444  &             0.9378 \\
\bottomrule
\end{tabular}%
}
\end{table*}

Table~\ref{tab.setupcomparison_firstlaymix} lists an extended architectural evaluation based on the top performing GBCNN configuration (\texttt{B-C-c-C-C-C}).
It reports the performance of the network configurations resulting from all possible combinations of receptive field and convolutional layers following the first layer.

We observe that the modification of the second layer has the largest impact on the amount of trainable parameters.
When using a second layer with SDPF filters and sparsity constraint (\texttt{B-B-c-C-C-C}), the amount of parameters is more than halved compared to using a fully convolutional one.
In accordance with the results from Table~\ref{tab: setupcomparison 5000}, receptive field layers impair the inpainting performance when placed in the image reconstruction section of the network (cf. Fig.~\ref{f.IRCNN_architecture}, layers four to six).
Out of those three positions, using SDPF filters in the last layer leads to the least degradation of results.

This yields two candidates with fewer parameters and only a slight degradation in performance compared to GBCNN: the configurations \texttt{B-C-c-C-C-B} with 170,001 parameters and \texttt{B-B-c-C-C-C} with 80,593 parameters.
We select \texttt{B-B-c-C-C-C} as the lightweight version of GBCNN and denote it as \emph{GBCNN\nobreakdash-L}.
Below, we will compare both network configurations against state-of-the-art methods.

\subsection{Filter analysis}

Here we analyze the properties of SDPF filters in the first network layer.

\begin{figure*}[tbh]
      \centering
      \begin{subfigure}[t]{\columnwidth}
      \centering
      \def\svgwidth{\columnwidth}
      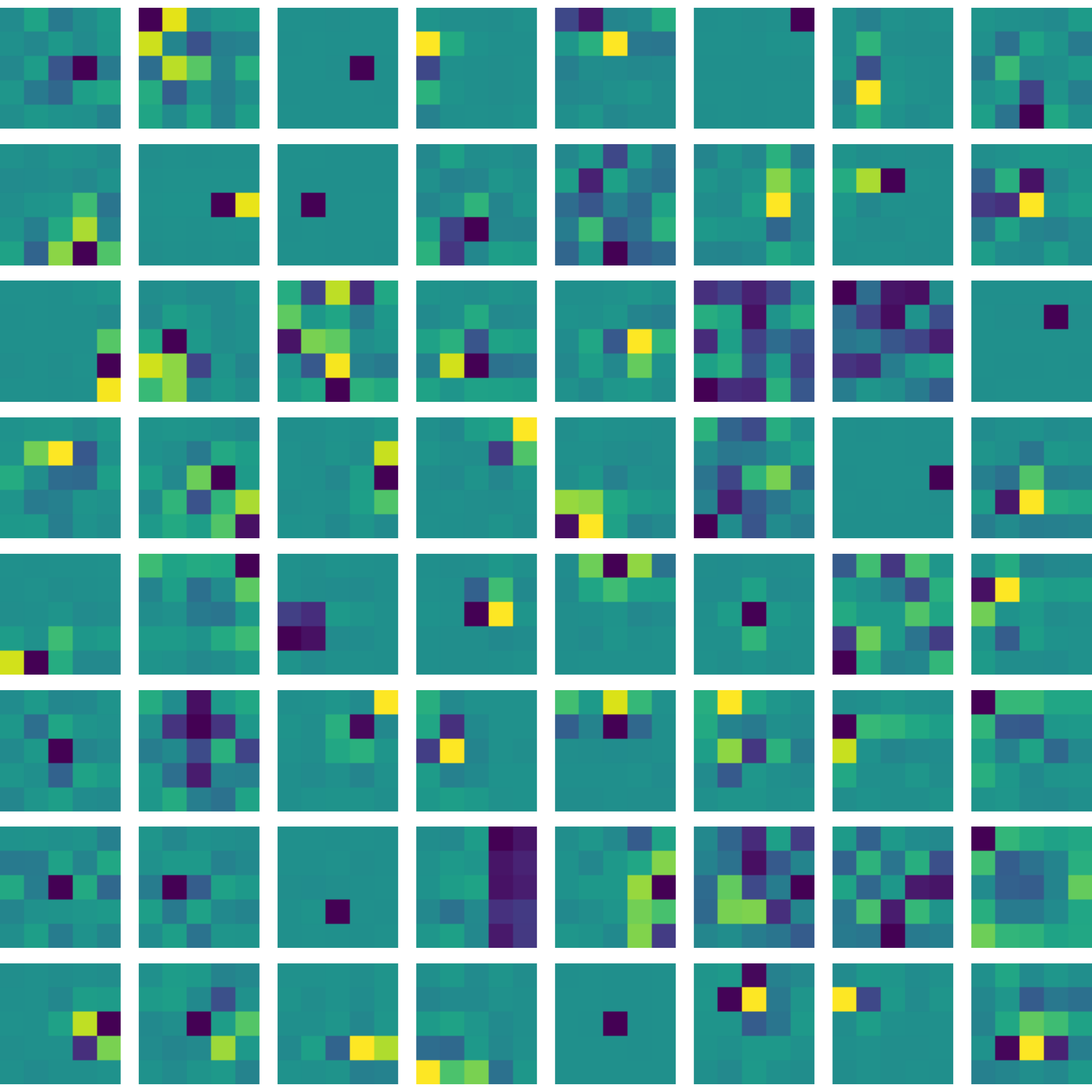
      \caption{Conventional convolutional kernels (layer type \texttt{C})}
      \label{f.trained-filters.IRCNN}
      \end{subfigure}\hfill
      ~
      \begin{subfigure}[t]{\columnwidth}
      \centering
      \def\svgwidth{\columnwidth}
      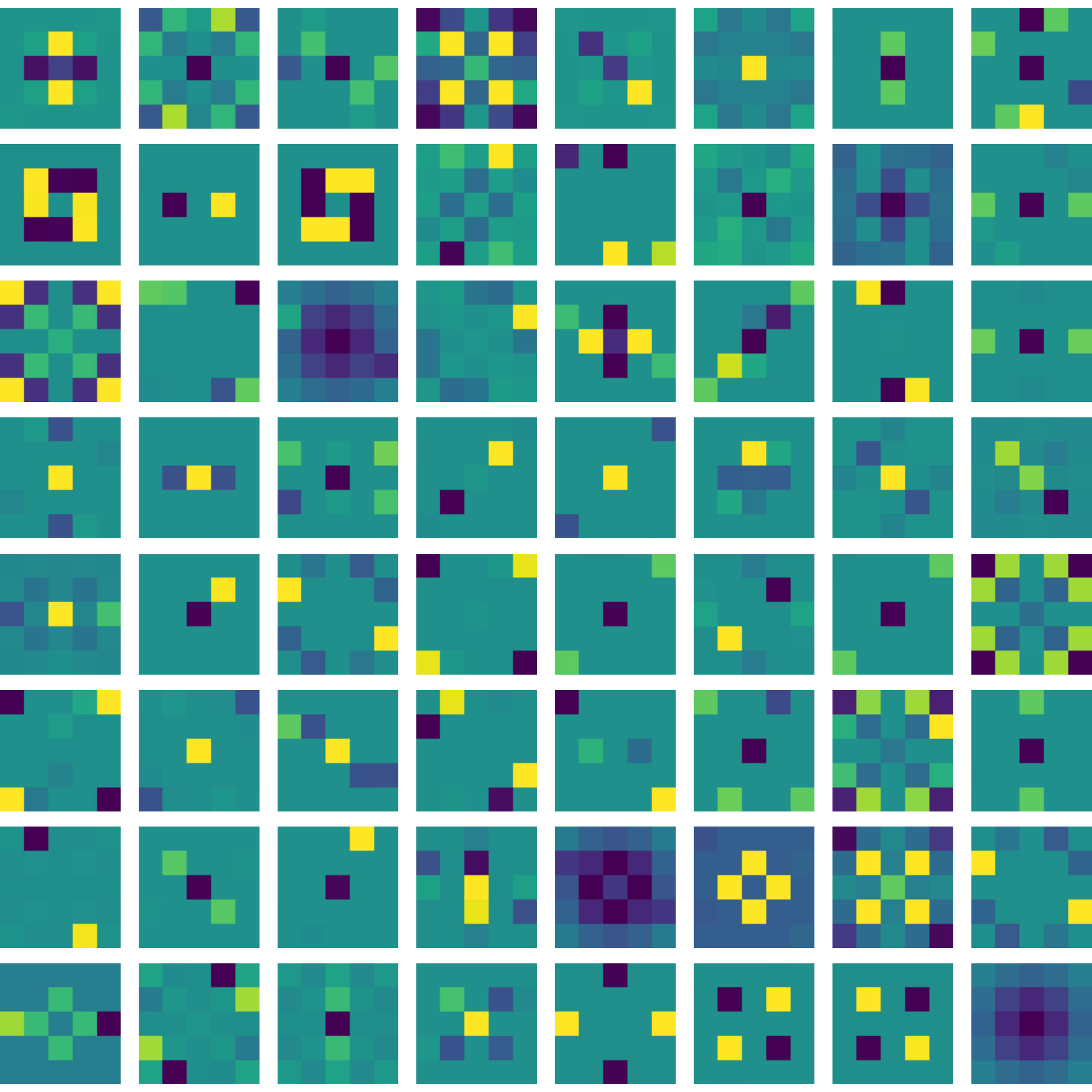
      \caption{Kernels in receptive field layer (layer type \texttt{B})}
      \label{f.trained-filters.GBCNN}
      \end{subfigure}
      \caption{The 64 learned filters in the first layer of the CNN architecture in Fig.~\ref{f.IRCNN_architecture} resulting from different training strategies.}
      \label{f.trained-filters}
\end{figure*}

\subsubsection{SDPF filter response}

To analyze the differences between convolutional and SDPF constrained receptive field layers, we investigate the structure of learned features when they are applied to natural images.
Fig.~\ref{f.trained-filters} visualizes the 64 filters in the first layer of IRCNN (\ref{f.trained-filters.IRCNN}) and GBCNN (\ref{f.trained-filters.GBCNN}) after being trained for 100 epochs on the full test set of 221,000 images.

Filters learned in the first layer of the GBCNN are visually more structured and include several first- and second order difference filters.
These clearly result from the linear combinations of a few elements from the SDPF dictionary.
We also notice a few low pass filters (e.g., row 2, column 7 and row 8, column 8).
By contrast, the IRCNN filters appear less symmetrical and include almost no elements identifiable as difference filters.

\begin{figure*}[bht]
      \centering
      \captionsetup[subfigure]{labelformat=empty}
      \begin{subfigure}[b]{.131\columnwidth}
      \centering
      \def\svgwidth{1\textwidth}
      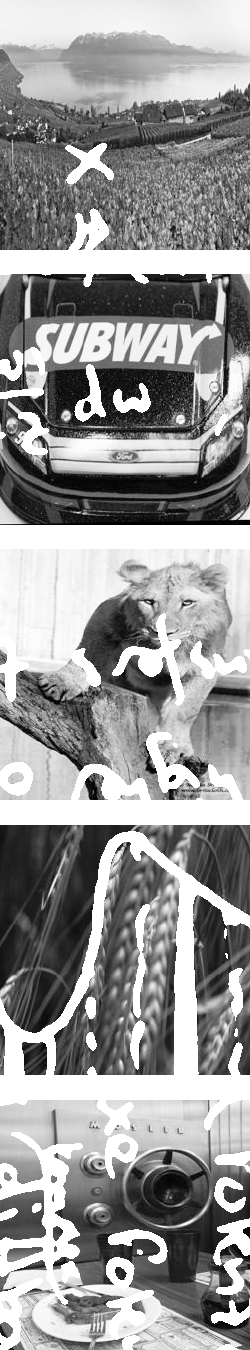
      \caption{\phantom{(a)}}
      \end{subfigure}
      \hspace{0.1\baselineskip}
      ~
      \captionsetup[subfigure]{labelformat=parens}
      \setcounter{subfigure}{0}
      \begin{subfigure}[b]{.85\columnwidth}
            \centering
            \def\svgwidth{1\textwidth}
            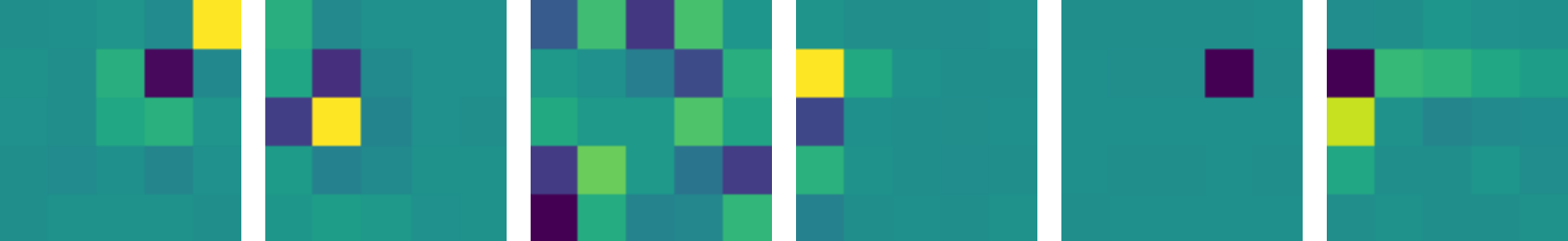
            \\
            \vspace{.8\baselineskip}
            \def\svgwidth{1\textwidth}
            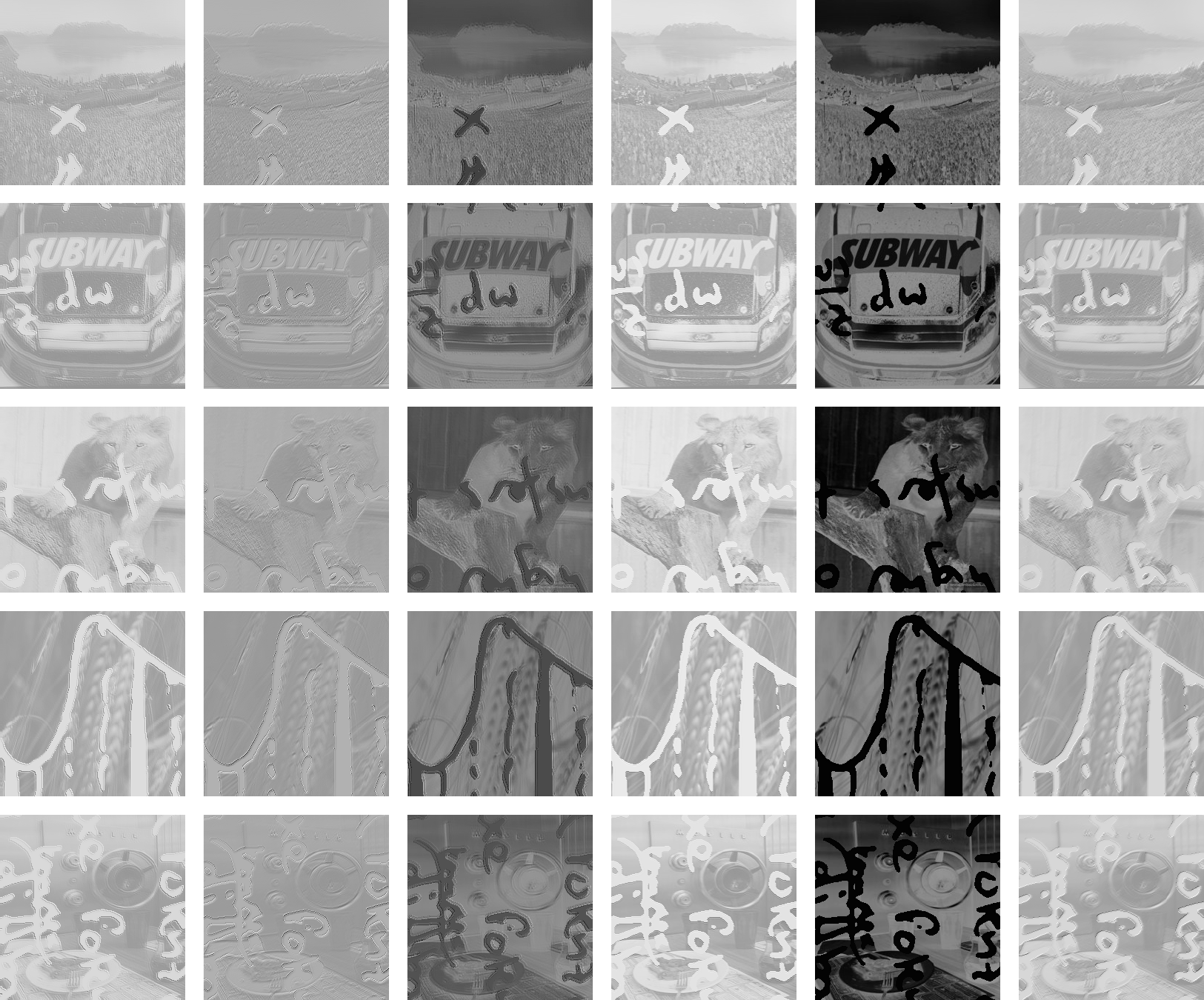
            \caption{IRCNN}
            \label{f.filtereffect.images.IRCNN}
      \end{subfigure}
      \hspace{0.1\baselineskip}
      ~
      \begin{subfigure}[b]{.85\columnwidth}
            \centering
            \def\svgwidth{1\textwidth}
            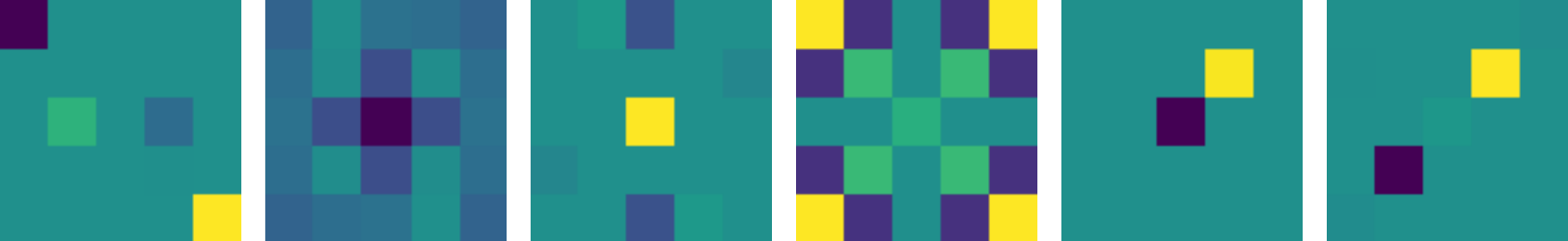
            \\
            \vspace{.8\baselineskip}
            \def\svgwidth{1\textwidth}
            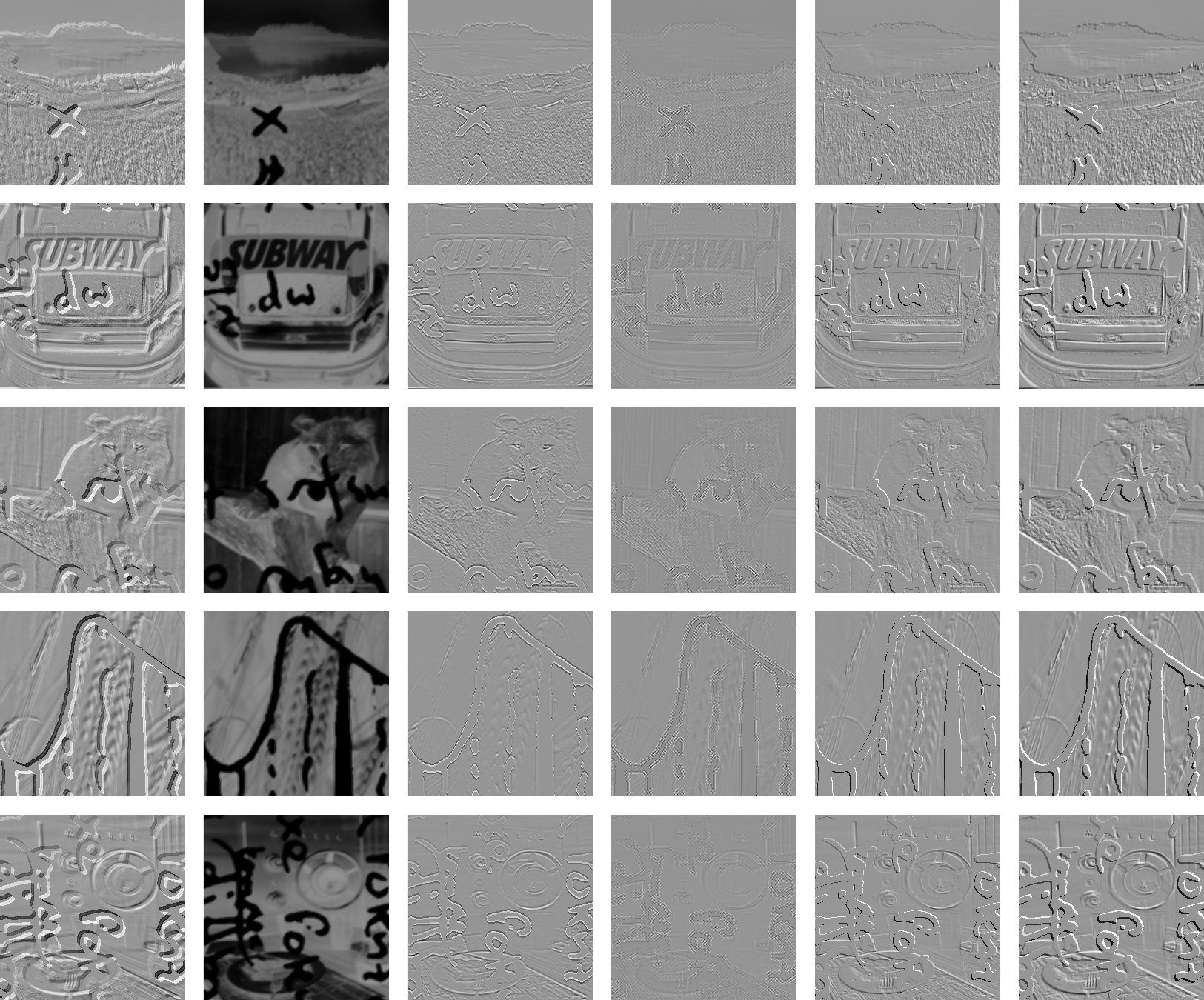
            \caption{GBCNN}
            \label{f.filtereffect.images.GBCNN}
      \end{subfigure}\hfill
      \caption{Filter responses on five natural images with overlaid handwriting for six randomly sampled filters from the first layer of the IRCNN and GBCNN networks (cf. Fig.~\ref{f.trained-filters} for entire filter sets). The filters are displayed in the top row and the original images are shown in the first column on the right.}
      \label{f.filtereffect.images}
      
      \vspace*{2\baselineskip}
%
      \def\svgwidth{.8\textwidth}
      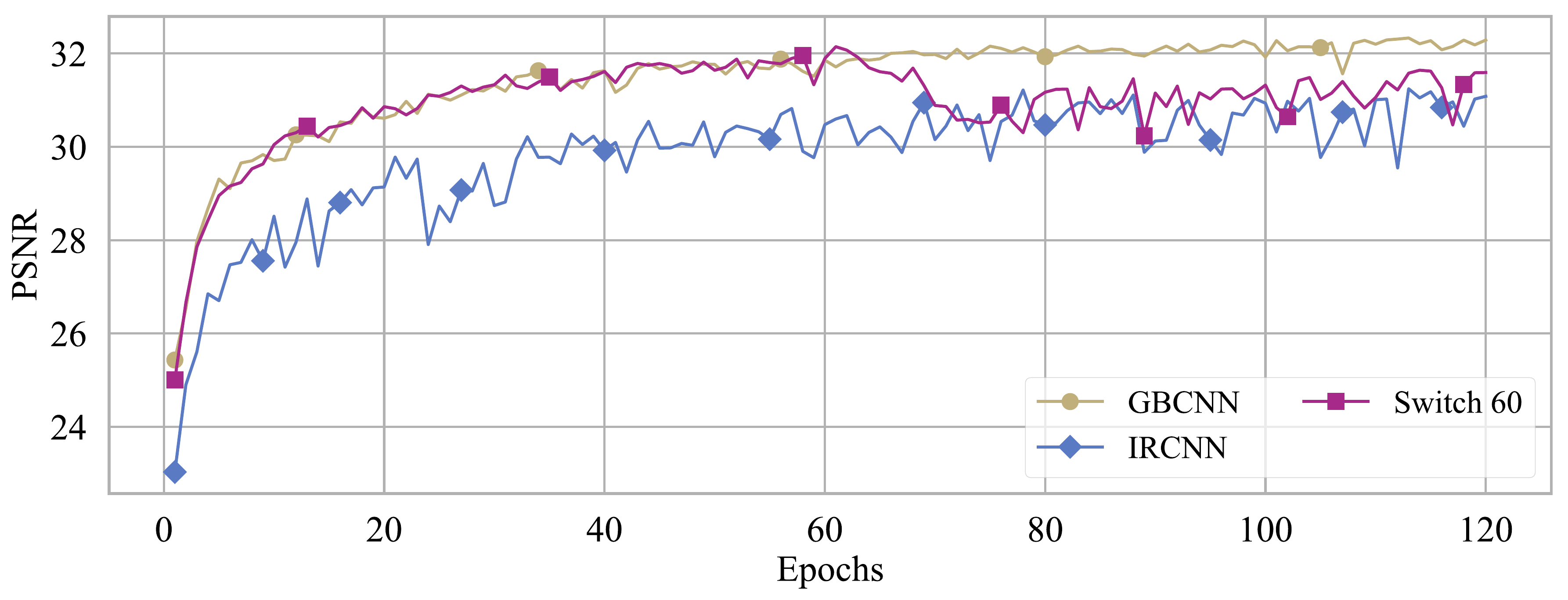
      \caption{Inpainting performance as a function of the number of training epochs. The GBCNN performance is always above the IRCNN performance. The purple curve (labeled \enquote{Switch 60}) shows the effect of relaxing the SDPF constrained receptive field layer into a conventional convolutional layer after 60 epochs; the network performance decreases to a  level comparable to the IRCNN. Results are averaged over 10 runs.}
      \label{f.switching}
\end{figure*}

In Fig.~\ref{f.filtereffect.images}, we illustrate the responses of a representative selection of filters applied to natural images, to highlight the different filter characteristics.
To this end, we visualize the effect of randomly selected filters from the first layers of IRCNN and GBCNN on images from the test data set.
The filters learned in the SDPF constrained receptive field  layers in Fig.~\ref{f.filtereffect.images.GBCNN} act typically as edge detectors along selected orientations (columns 1,3,5,6) but also include a low-pass (column 2) and high pass (column 4) filter.
By contrast, the filter responses for the IRCNN architecture in Fig.~\ref{f.filtereffect.images.IRCNN} are less structured and do not include strong directional responses.
This highlights the geometric character and interpretability of the filters associated with our SDPF constrained receptive field layers that are inspired by principles of sparse image approximations.

\subsubsection{SDPF filters during training}

Here we investigate whether applying our strategy based on pre-designed filters is advantageous throughout the whole training process.
One could suppose that these filters simply act as a good \enquote{layer initialization} during the first epochs; one could then lift the SDPF dictionary constraint after a certain number of epochs after which the network would convergence to an even better model.
To investigate this possibility, we implemented a version of a receptive field layer that we trained for a fixed number of epochs under the constraint that convolutional kernels are taken as linear combinations our SDPF dictionary elements; after a prescribed number of epochs, we used the learned filters as initial weights for a conventional convolutional layer that we trained further.

Fig.~\ref{f.switching} shows the result of this numerical experiment, where we trained a GBCNN for 60 epochs, after which we lifted the SDPF constraint.
This effectively replaces the SDPF constrained receptive layer with a conventional convolutional layer that we initialized with the filters learned after 60 epochs.
The figure compares the inpainting performance of the model resulting from this experiment against a GBCNN and an IRCNN.
Remarkably, despite the good initialization after 60 epochs, the performance of the network trained with conventional convolutional layers decreases with respect to the GBCNN and falls to the level of a network that used convolutions during the whole training.

\FloatBarrier

\begin{figure}[thb]
      \centering
      \def\svgwidth{0.95\columnwidth}
      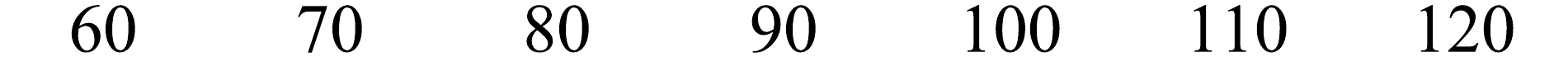
      \vspace*{.5\baselineskip}
      \def\svgwidth{0.95\columnwidth}
      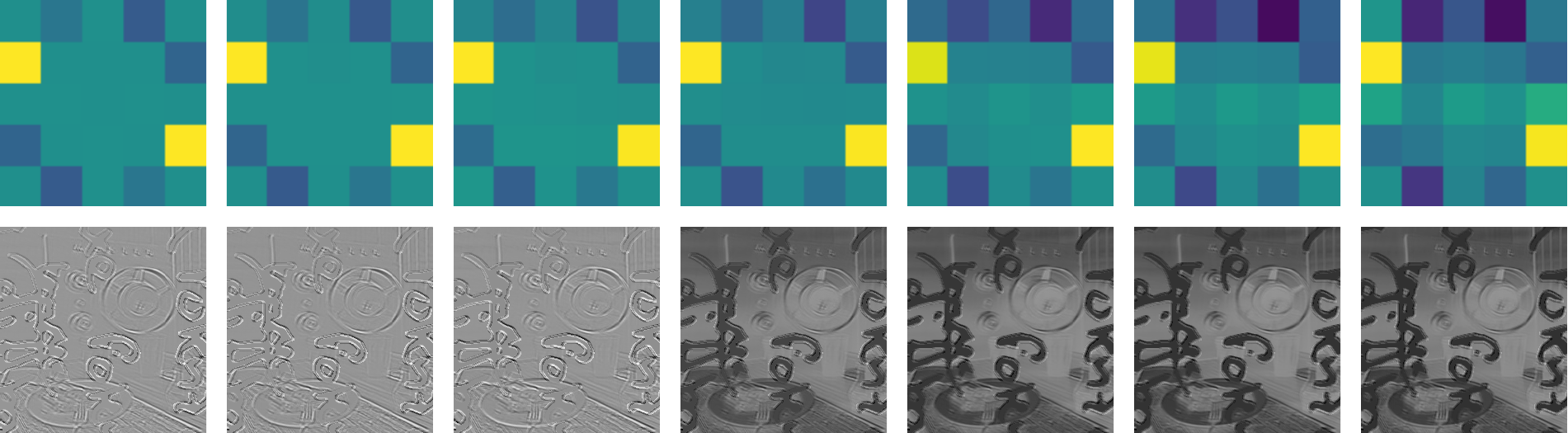
      \vspace*{.5\baselineskip}
      \def\svgwidth{0.95\columnwidth}
      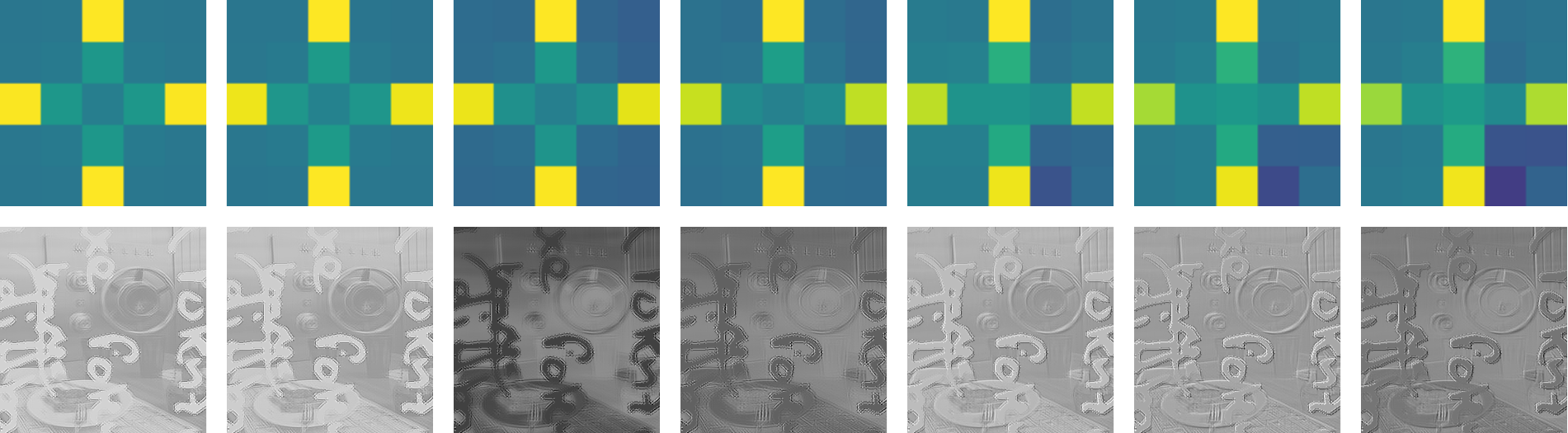
      \vspace*{.5\baselineskip}
      \def\svgwidth{0.95\columnwidth}
      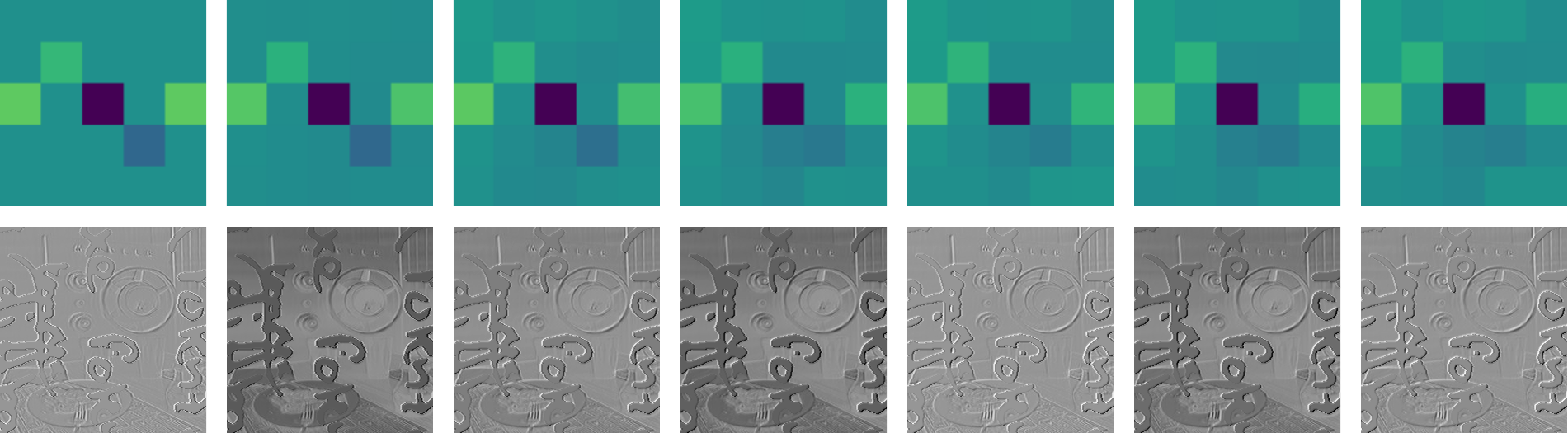
      \vspace*{.5\baselineskip}
      \def\svgwidth{0.95\columnwidth}
      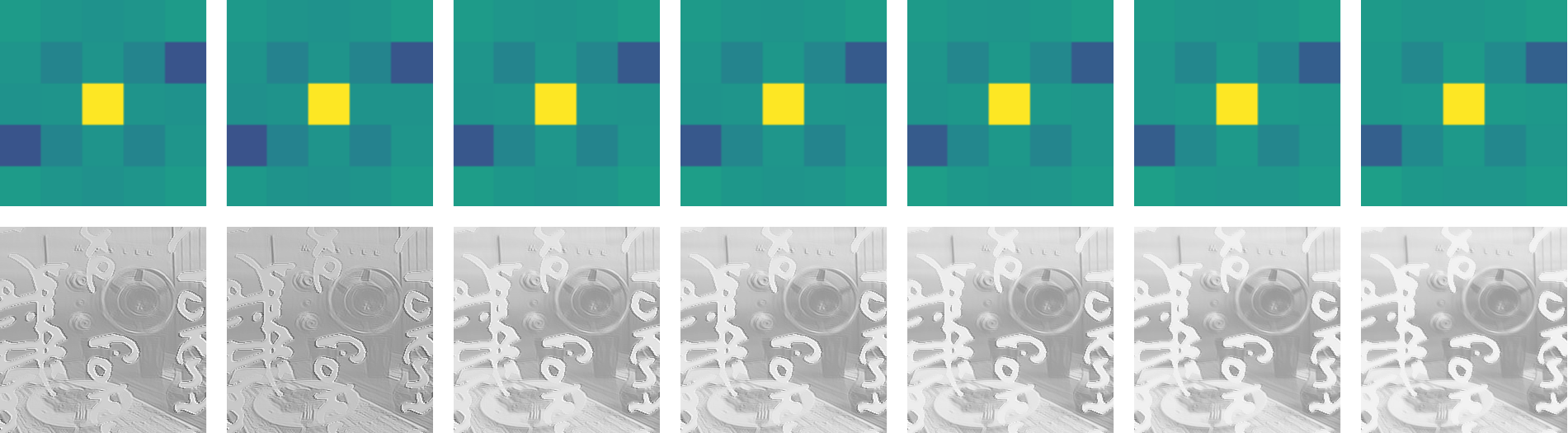
      \vspace*{.5\baselineskip}
      \def\svgwidth{0.95\columnwidth}
      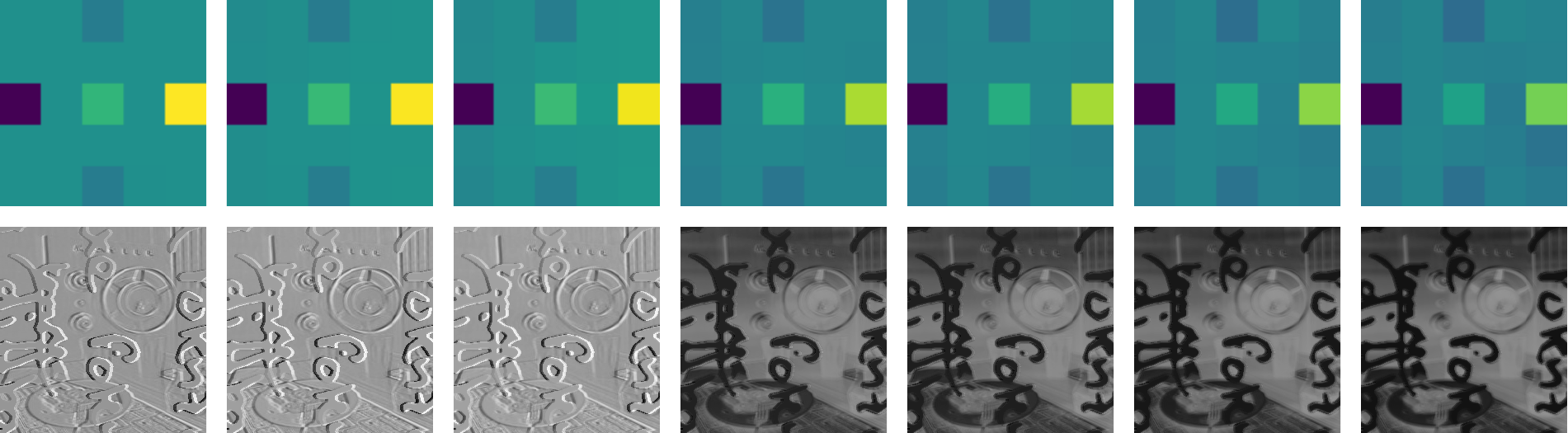
      \def\svgwidth{0.95\columnwidth}
      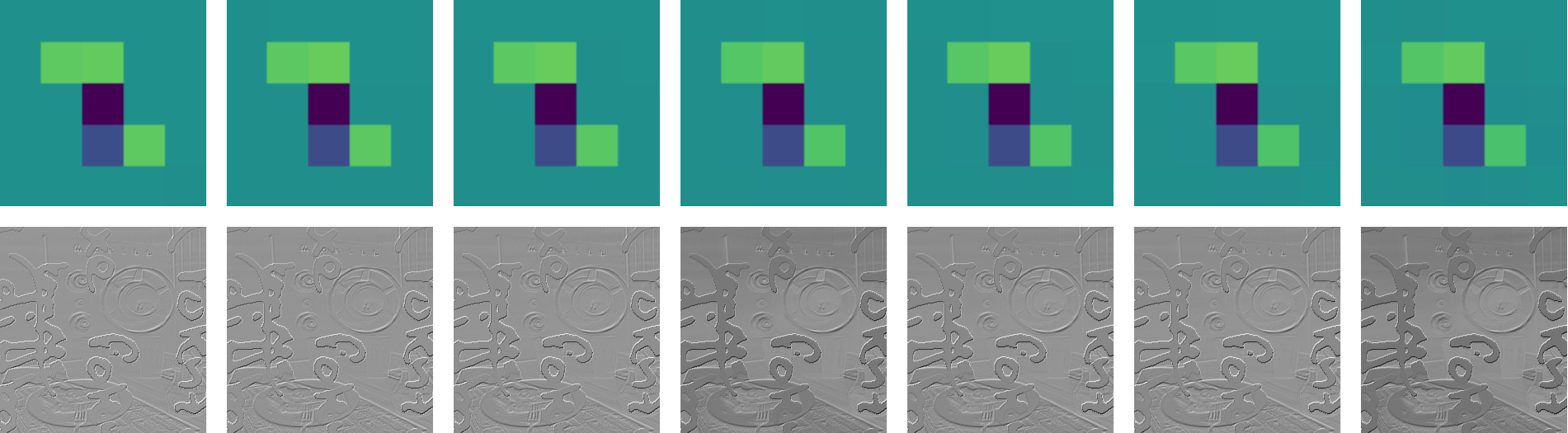
      \caption{Evolution of selected kernels in the first network layer, after the receptive field layer is relaxed into a conventional one. Each kernel is displayed above its response every 10 epochs, between 60-120 epochs.}
      \label{f.filtereffect}
\end{figure}

\FloatBarrier

\begin{figure*}[thb]
      \centering
      \def\svgwidth{.8\textwidth}
      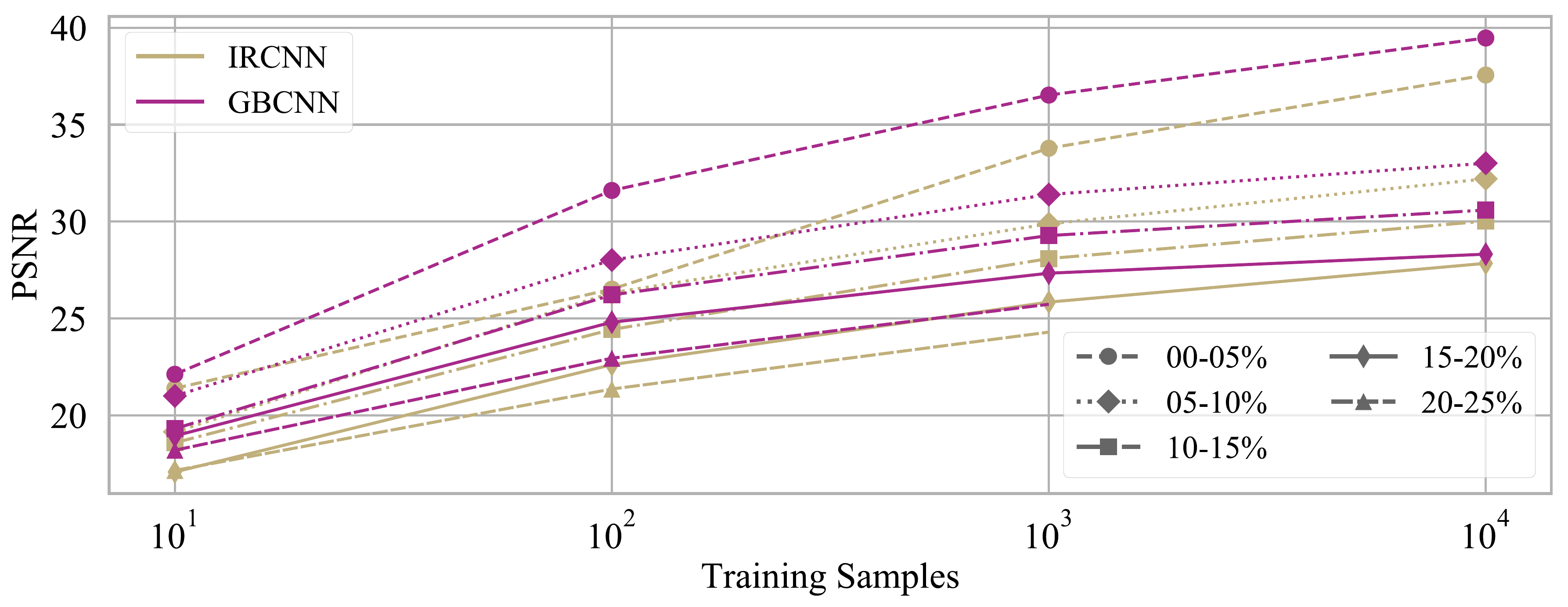
      \caption{Image inpainting performance (PSNR) of IRCNN and GBCNN, for different occlusion areas (distinguished by line styles) on the training images. For 20-25\% coverage only $10^3$ samples were available. Networks are evaluated after training for $100$ epochs and the numbers are averaged over $10$ network trainings.}
      \label{f.psnr2}
\end{figure*}

To further illustrate this phenomenon, Fig.~\ref{f.filtereffect} visualizes a randomly selected set of filters over the duration between lifting the SDPF constraint until the training ends.
We see that the filter responses change dramatically, even though the filters appear to change very gradually.
In some instances, the filter responses lose their sensitivity to edges and salient features that is common for filters obtained from the SDPF dictionary (cf.~Fig.~\ref{f.filtereffect.images.IRCNN})

All in all, these observations show that our training strategy based on SPDF dictionary and sparsity constraint affects the convergence of the network kernels throughout the entire training process and its effect cannot be reduced to a clever filter initialization.

\subsection{Inpainting performance with increasing training set size}

In practical applications of neural networks, it is often important to decide the amount of training data that is required for the network to converge to a satisfactory model.

We systematically investigated the inpainting performance of our GBCNN as compared to a IRCNN for different amounts of training data and reported the results in Fig.~\ref{f.psnr2}.
The figure shows the image inpainting performance in PSNR for images corrupted by occlusions affecting a different fraction of the image area as a function of the number of training samples (between 10 and 10,000 samples).
The networks are trained for~$100$ epochs and the results averaged over 10 runs.
Compared to IRCNN, our GBCNN approach exhibits higher PSNR values on the test set after being trained on relatively few data samples.
This difference in performance reduces with more data, but remains very significant on images with smaller areas to inpaint.

Since our GBCNN  uses a lower number of trainable parameters than an IRCNN (with the same architecture), it was expected that the former would converge with a lower number of training samples.
This intuition was confirmed by our numerical experiments.
We explain this behavior with the improved ability of the SPDF dictionary to capture the essential image characteristics which is also the reason for the competitive performance found in the shearlet-based inpainting algorithm that motivated this study.
By contrast, the IRCNN requires more training samples to reach the same performance as the GBCNN.

\subsection{Benchmark comparison of inpainting performance}

We compared the performance of our GBCNN and GBCNN\nobreakdash-L against state-of-the-art algorithms for blind image inpainting, including IRCNN~\cite{chaudhury_can_2017} and VCNet~\cite{wang2020vcnet}.
Our comparison does not include the wide body of work on non-blind deep inpainting methods, e.g., partial convolutions~\cite{liu_image_2018}, contextual attention~\cite{Yu_2018_CVPR}, convolutions~\cite{Yu_2019_ICCV} or generative approaches, due to their use of information about the corruption's location and the resulting problem simplification.
However, we included the non-blind method ShearLab~\cite{king_analysis_2014} in the comparison, as our approach is partially motivated by sparsity-based ideas underlying this approach.

All networks are trained on the full data set ($221,000$ images) for 100 epochs, except for VCNet which is trained from scratch for 160,000 epochs\footnote{Following the original paper's training schedule. This became necessary, since the pre-trained VCNet model provided by the authors in~\cite{wang2020vcnet} did not achieve comparable results even after tuning it on our data set for 100,000 additional epochs.}.
Each network is trained five times, and the training run with the best performance is displayed.

\begin{table*}[h!]
   \center
   \caption{Image inpainting performance measured as average PSNR on the entire test set or on subsets associated with the percentage of image lost due to occlusion. Best result by column in bold.}
   \label{tab.comparison}
   \scalebox{0.85}{
   \begin{tabular}{@{\ }lrrrrrr@{\ }}
    \toprule
    \textbf{Method} &   \textbf{entire set} &  \textbf{0--5\%}   &  \textbf{5--10\%}  &   \textbf{10--15\%}  &  \textbf{15--20\%} & \textbf{20--25\%}  \\
    \midrule
    Shearlet~\cite{king_analysis_2014} &           30.7005  &           34.5694&             31.4638&            29.5394&                    27.7532  &                25.4599 \\
    \midrule
    VCNet~\cite{wang2020vcnet}         &         32.6120    &         37.8666  &           33.1650  &  \textbf{30.9287} &            \textbf{29.0408} &      \textbf{27.0796} \\
    IRCNN~\cite{chaudhury_can_2017}    &         32.5554    &         38.9941  &           33.0309  &          30.5410  &                    28.2957  &              26.1550 \\
    GBCNN (ours)                       &         32.9497    &         39.5763  &   \textbf{33.3030} &          30.8849  &                    28.6630  &              26.6679 \\
    GBCNN\nobreakdash-L  (ours)        & \textbf{33.0063}   & \textbf{39.9796} &           33.2698  &          30.8267  &                    28.5877  &              26.6221 \\
    \bottomrule
    \end{tabular}
   }
\end{table*}

\begin{table*}[htb]
   \center
   \caption{Comparison of network parameter count and average evaluation time $\mu_t$ per image (size $256\times256$) in milliseconds for inpainting methods.}%
   \label{tab: evaluation time comparison}
   \scalebox{0.85}{
   \begin{tabular}{@{\ }lrrrrr@{\ }}
    \toprule
                &   \textbf{ShearLab}~\cite{king_analysis_2014} & \textbf{VCNet}~\cite{wang2020vcnet} & \textbf{IRCNN}~\cite{chaudhury_can_2017} & \textbf{GBCNN} & \textbf{GBCNN\nobreakdash-L}\\
     \midrule
     Parameters & -- & 3,789,892 & 172,113 & 170,705 & \textbf{80,593} \\
     \midrule
     $\mu_t$ [msec] & 29,147.001 & 17.460 & 2.539 & \textbf{2.453} & 2.534 \\
    \bottomrule
    \end{tabular}
   }
\end{table*}

\begin{figure*}[thb]
\centering
\begin{tabular}{@{}c@{\ }c@{\ \ \ \ \ }c@{\ }c@{\ }c@{\ }c@{\ }c@{}}
      & & \small 28.77 & \small 29.89 & \small 30.24 & \small 30.46 & \small \textbf{30.53}\\
      \includegraphics[width=0.121\textwidth]{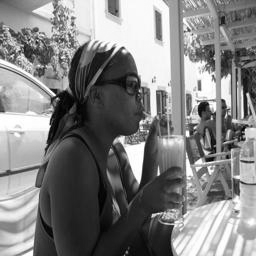} &
      \includegraphics[width=0.121\textwidth]{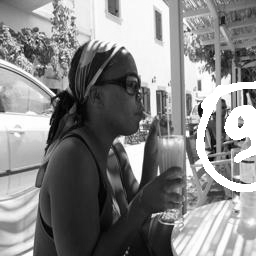} &
      \includegraphics[width=0.121\textwidth]{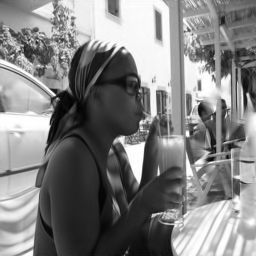} &
      \includegraphics[width=0.121\textwidth]{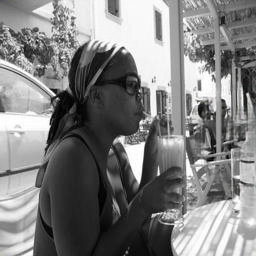} &
      \includegraphics[width=0.121\textwidth]{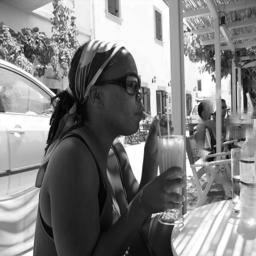} &
      \includegraphics[width=0.121\textwidth]{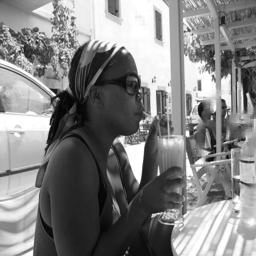} &
      \includegraphics[width=0.121\textwidth]{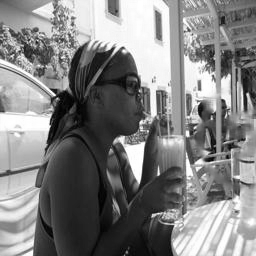} \\
      & & \small 32.80 & \small 34.63 & \small 34.68 & \small 35.12 & \small \textbf{35.18}\\
      \includegraphics[width=0.121\textwidth]{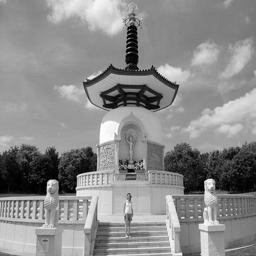} &
      \includegraphics[width=0.121\textwidth]{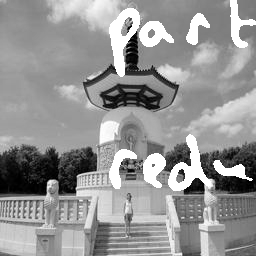} &
      \includegraphics[width=0.121\textwidth]{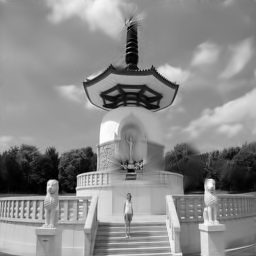} &
      \includegraphics[width=0.121\textwidth]{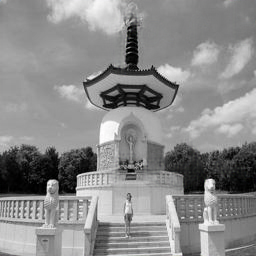} &
      \includegraphics[width=0.121\textwidth]{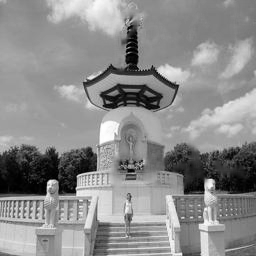} &
      \includegraphics[width=0.121\textwidth]{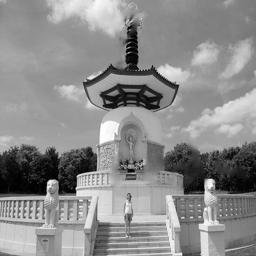} &
      \includegraphics[width=0.121\textwidth]{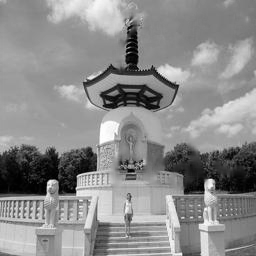} \\
      & & \small 29.73 & \small 30.65 & \small 31.07 & \small 31.46 & \small \textbf{31.65}\\
      \includegraphics[width=0.121\textwidth]{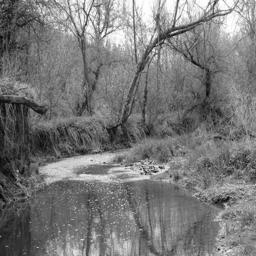} &
      \includegraphics[width=0.121\textwidth]{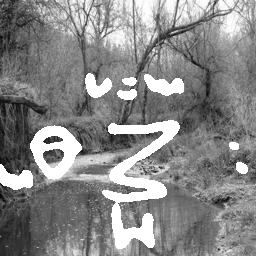} &
      \includegraphics[width=0.121\textwidth]{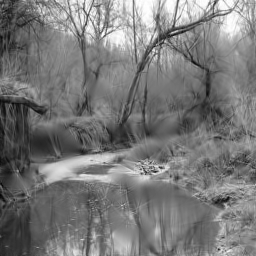} &
      \includegraphics[width=0.121\textwidth]{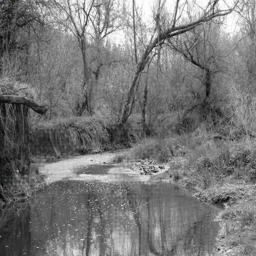} &
      \includegraphics[width=0.121\textwidth]{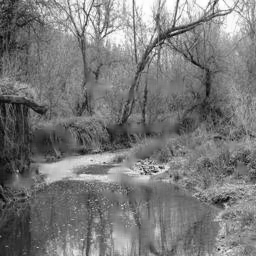} &
      \includegraphics[width=0.121\textwidth]{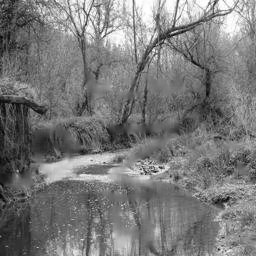} &
      \includegraphics[width=0.121\textwidth]{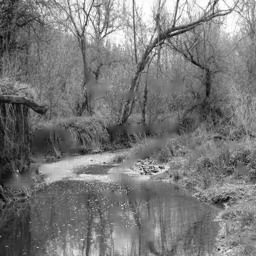} \\
      & & \small 28.09 & \small 28.53 & \small 28.80 & \small \textbf{29.14} & \small 28.98\\
      \includegraphics[width=0.121\textwidth]{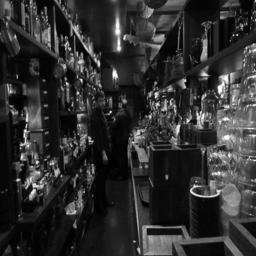} &
      \includegraphics[width=0.121\textwidth]{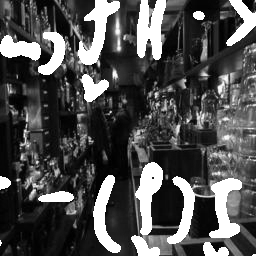} &
      \includegraphics[width=0.121\textwidth]{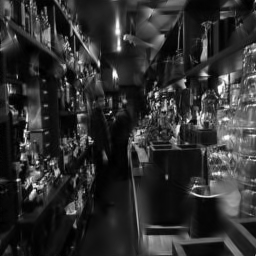} &
      \includegraphics[width=0.121\textwidth]{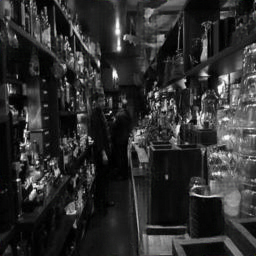} &
      \includegraphics[width=0.121\textwidth]{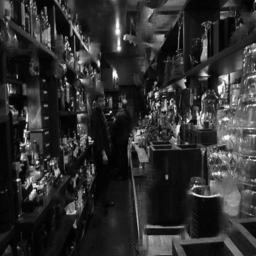} &
      \includegraphics[width=0.121\textwidth]{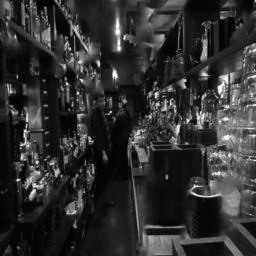} &
      \includegraphics[width=0.121\textwidth]{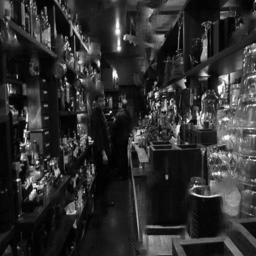} \\
      & & \small 30.57 & \small \textbf{30.70} & \small 29.76 & \small 30.66 & \small 30.66\\
      \includegraphics[width=0.121\textwidth]{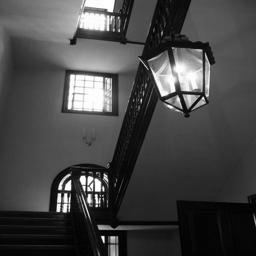} &
      \includegraphics[width=0.121\textwidth]{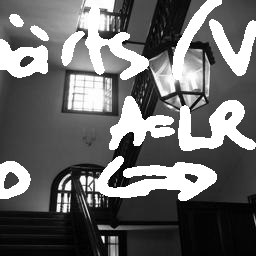} &
      \includegraphics[width=0.121\textwidth]{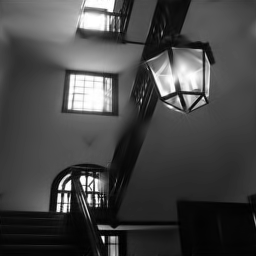} &
      \includegraphics[width=0.121\textwidth]{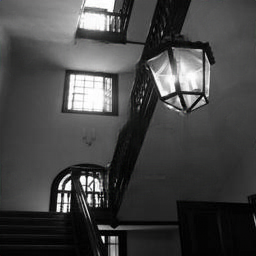} &
      \includegraphics[width=0.121\textwidth]{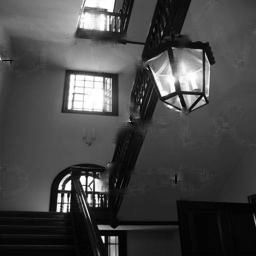} &
      \includegraphics[width=0.121\textwidth]{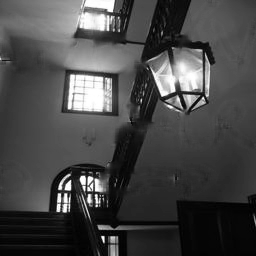} &
      \includegraphics[width=0.121\textwidth]{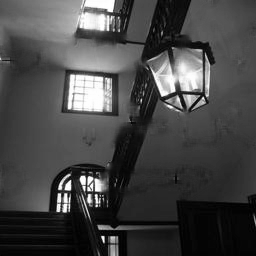} \\
      & & \small 28.68 & \small 29.76 & \small 29.77 & \small \textbf{30.16} & \small 29.87\\
      \includegraphics[width=0.121\textwidth]{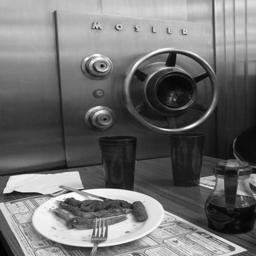} &
      \includegraphics[width=0.121\textwidth]{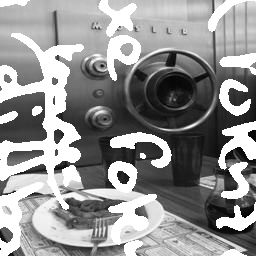} &
      \includegraphics[width=0.121\textwidth]{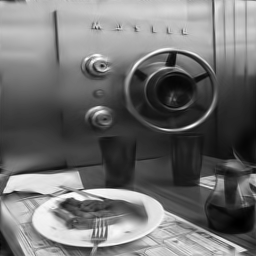} &
      \includegraphics[width=0.121\textwidth]{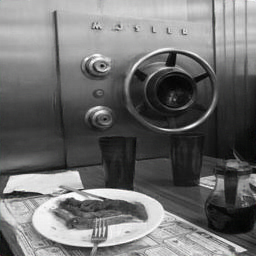} &
      \includegraphics[width=0.121\textwidth]{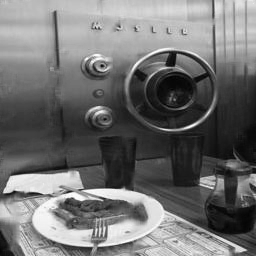} &
      \includegraphics[width=0.121\textwidth]{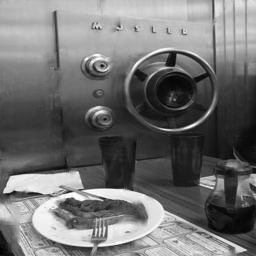} &
      \includegraphics[width=0.121\textwidth]{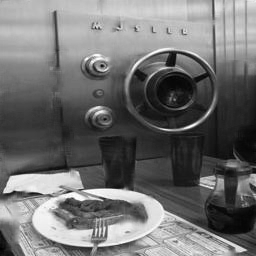} \\
      & & \small 21.78 & \small \textbf{26.69} & \small 22.74 & \small 23.63 & \small 23.58\\
      \includegraphics[width=0.121\textwidth]{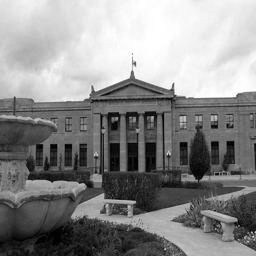} &
      \includegraphics[width=0.121\textwidth]{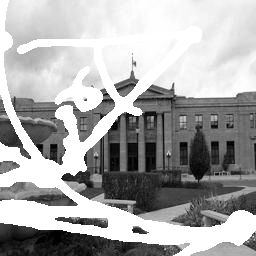} &
      \includegraphics[width=0.121\textwidth]{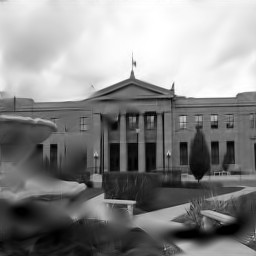} &
      \includegraphics[width=0.121\textwidth]{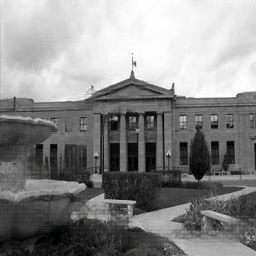} &
      \includegraphics[width=0.121\textwidth]{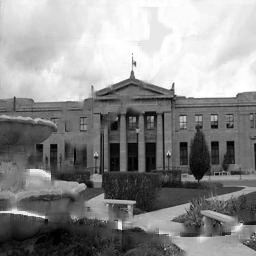} &
      \includegraphics[width=0.121\textwidth]{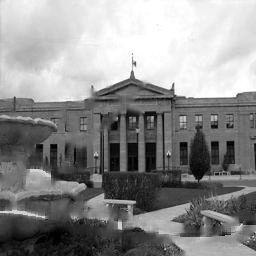} &
      \includegraphics[width=0.121\textwidth]{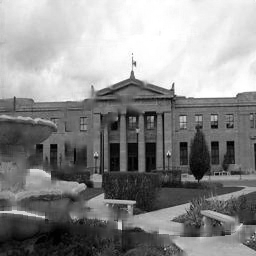} \\
      \small Original &  Masked & Shearlet~\cite{king_analysis_2014} & VCNet\cite{wang2020vcnet} & IRCNN~\cite{chaudhury_can_2017} & GBCNN & GBCNN\nobreakdash-L\\
\end{tabular}
\caption{Visual comparison of inpainting methods on images from different occlusion percentages; PSNR for each method is reported above the restored image.}\label{fig.comparison}
\end{figure*}

Table~\ref{tab.comparison} reports the image inpainting performance in PSNR on the full test set and on its subsets associated with different fractions of the image area lost to occlusion.
Since the qualitative performance, as measured using the SSIM metric, is very similar to the PSNR, we did not report it.

The table shows that, on the entire test set, our lightweight GBCNN-L architecture outperforms all tested methods, with the GBCNN performing very closely; the more sophisticated VCNet performs worse than GBCNN but better than IRCNN.
When we examine the performance on each subset, the table shows that GBCNN\nobreakdash-L outperforms the other methods  on images with relatively small occlusions (0-5\%), with GBCNN performing closely and VCNet showing a significantly worse performance (37.8666 vs 39.9796 dBs).
For larger occlusions (5-10\%) GBCNN has the best performance with GBCNN\nobreakdash-L performing closely.
For the largest tested occlusions (15-20\% and 20-25\%) VCNet achieves the best performance, even though the improvement with respect to GBCNN is less that 0.5 dBs.

The table shows that all learning-based approaches consistently outperform the (non-blind) shearlet-based inpainting method from~\cite{king_analysis_2014}.
This is remarkable, considering that the latter method is non-blind, and confirms the superior performance of learning-based methods in inpainting.

The comparison of the different network approaches shows that, due to their ability to capture salient image features in images, our  GBCNN and GBCNN\nobreakdash-L perform very competitively overall, even outperforming the more sophisticated VCNet approach on the full data set.
The performance of our method is particularly impressive on images where a relatively small fraction of the area is affected by occlusion.
As expected, when the  area to inpaint increases in size, the information about the location of the missing region becomes more important and VCNet, taking advantage of its greater complexity, is able to recover large blocks of missing image information more effectively.
Yet, the performance of GBCNN is not far off (about -0.4dBs).

We remark that VCNet requires $1,600$ times more training epochs than GBCNN and uses a complex multistep algorithm.
We also remark that GBCNN and GBCNN\nobreakdash-L apply a $5\times 5$ SDPF dictionary, which limits the support size of the edge-like elements that these methods are able to capture (and recover) with high efficiency.

Representative reconstruction results for all inpainting methods are shown in Fig.~\ref{fig.comparison}.
In terms of visual quality, close inspection shows that our approach is less affected by blur as compared to VCNet.
We observed this property consistently and interpret it as a consequence of our special filter selection process that is designed to capturing edge-like structures with high efficiency.
This observation could also explain the worse performance of VCNet on images where only a relatively small fraction of the area is affected by occlusion, as loosing high frequency content would degrade the reconstruction quality.

\subsubsection{Evaluation time and parameters}

We also compared the inference times for all of the inpainting algorithms considered in this study in Table~\ref{tab: evaluation time comparison}, and additionally considered the number of trainable parameters.
All run times are computed using  a single NVIDIA Tesla V100 PCIe graphics card with 32 GB memory.
All network approaches were implemented in Python and Tensorflow, while the shearlet-based method has a Matlab implementation.

Top times are delivered by IRCNN, GBCNN and GBCNN\nobreakdash-L with about 2.5 msec, since inference requires a computationally cheap feed forward pass on a simple network architecture.
The time differences between these three methods are negligible, and might be due to small fluctuations in the time measurements.
Due to the lightweight architecture resulting from the placement of receptive field layers, GBCNN\nobreakdash-L has the lowest parameter count, followed by GBCNN and IRCNN with twice as many parameters.
Compared to the three networks above, VCNet has about 20 times more parameters and takes roughly 5 times longer for inference.
This comparison shows the advantages of SDPF constrained receptive field layers for the development of lightweight and fast network architectures.
The iterative nature of the shearlet-based inpainting algorithm significantly increases its run time (about $10^4$ times larger) as compared to all network implementations.

\section{Conclusions}

We have introduced a novel strategy for blind image inpainting that brings model-based principles from the theory of sparse representation into the design of a new deep learning model.
Our approach employs a specifically designed filter dictionary, called SPDF, in combination with a sparsity constraint, that is motivated by the success of shearlet representations in image processing applications.
With these novel concepts we develop two lightweight network models called GBCNN and GBCNN\nobreakdash-L.

One main advantage of our approach to blind inpainting is the increased interpretability.
As compared to the conventional CNN approach where there is essentially no control on the kernel structure, the kernels learned by our GBCNN and GBCNN\nobreakdash-L reflect the geometric properties of the SPDF dictionary that are critical to capture salient image features such as edges and corners.
This behavior is consistent with the model-based principles that guided our design strategy.

By integrating model-based principles into a simple and light weight network architecture, our approach outperforms not only conventional CNN schemes with similar architectures in terms of inpainting quality.
It also excels the significantly more complex VCNet algorithm, which applies a multistep strategy for blind image inpainting.

However, our approaches exhibit a reduced inpainting quality in the case where the region to be inpainted is relatively large.
We believe the reduction in competitiveness of our method in this case to be explained by the support size of our filter dictionary, whose element were selected to have fixed size of $5 \times 5$ pixels.
This observation suggests that our method could possibly be improved by considering a filter dictionary with elements having multiple size supports, e.g., $5 \times 5$, $7 \times 7$ and $9 \times 9$.
This extension would be fully consistent with the theoretical framework of shearlet-based inpainting \cite{guo2020} that inspired the present work, since the desirable properties of the shearlet representation system include not only directional sensitivity but also multiresolution.
We recall that the idea of using filters of different support size in CNNs has been already employed successfully in the deep learning literature, most notably in the celebrated Inception architecture \cite{szegedy2015}.

\FloatBarrier

\section*{Acknowledgments}

Authors thank Prof. Dr. D. Göddeke for enabling this collaboration and his constant support and advise, Prof. Dr. B. Haasdonk and Prof. Dr. C. Rohde for providing the handwritten notes, and K. Safari for his help with the implementation.
DL acknowledges support by NSF-DMS 1720487, 1720452 and HPE DSI/IT at UH.
JS acknowledges support by the Deutsche Forschungsgemeinschaft (DFG, German Research Foundation) - Project-ID 251654672 - TRR 161 (Project B04) at US and by the International Max Planck Research School for Intelligent Systems (IMPRS-IS).

\bibliographystyle{abbrv}
\bibliography{bibliography}
\end{document}

%% file: Images/5x5_new_parseval_visualization_2_49_viridis.pdf_tex
\begingroup%
  \makeatletter%
  \providecommand\color[2][]{%
    \errmessage{(Inkscape) Color is used for the text in Inkscape, but the package 'color.sty' is not loaded}%
    \renewcommand\color[2][]{}%
  }%
  \providecommand\transparent[1]{%
    \errmessage{(Inkscape) Transparency is used (non-zero) for the text in Inkscape, but the package 'transparent.sty' is not loaded}%
    \renewcommand\transparent[1]{}%
  }%
  \providecommand\rotatebox[2]{#2}%
  \newcommand*\fsize{\dimexpr\f@size pt\relax}%
  \newcommand*\lineheight[1]{\fontsize{\fsize}{#1\fsize}\selectfont}%
  \ifx\svgwidth\undefined%
    \setlength{\unitlength}{504bp}%
    \ifx\svgscale\undefined%
      \relax%
    \else%
      \setlength{\unitlength}{\unitlength * \real{\svgscale}}%
    \fi%
  \else%
    \setlength{\unitlength}{\svgwidth}%
  \fi%
  \global\let\svgwidth\undefined%
  \global\let\svgscale\undefined%
  \makeatother%
  \begin{picture}(1,1)%
    \lineheight{1}%
    \setlength\tabcolsep{0pt}%
    \put(0,0){\includegraphics[width=\unitlength,page=1]{5x5_new_parseval_visualization_2_49_viridis.pdf}}%
  \end{picture}%
\endgroup%

%% file: Images/NW-Struct_public.pdf_tex
\begingroup%
  \makeatletter%
  \providecommand\color[2][]{%
    \errmessage{(Inkscape) Color is used for the text in Inkscape, but the package 'color.sty' is not loaded}%
    \renewcommand\color[2][]{}%
  }%
  \providecommand\transparent[1]{%
    \errmessage{(Inkscape) Transparency is used (non-zero) for the text in Inkscape, but the package 'transparent.sty' is not loaded}%
    \renewcommand\transparent[1]{}%
  }%
  \providecommand\rotatebox[2]{#2}%
  \newcommand*\fsize{\dimexpr\f@size pt\relax}%
  \newcommand*\lineheight[1]{\fontsize{\fsize}{#1\fsize}\selectfont}%
  \ifx\svgwidth\undefined%
    \setlength{\unitlength}{577.57203034bp}%
    \ifx\svgscale\undefined%
      \relax%
    \else%
      \setlength{\unitlength}{\unitlength * \real{\svgscale}}%
    \fi%
  \else%
    \setlength{\unitlength}{\svgwidth}%
  \fi%
  \global\let\svgwidth\undefined%
  \global\let\svgscale\undefined%
  \makeatother%
  \begin{picture}(1,0.35788853)%
    \lineheight{1}%
    \setlength\tabcolsep{0pt}%
    \put(0,0){\includegraphics[width=\unitlength,page=1]{NW-Struct_public.pdf}}%
    \put(0.1272425,0.09079295){\color[rgb]{0,0,0}\makebox(0,0)[lt]{\lineheight{1.25}\smash{\begin{tabular}[t]{l}$5 \times 5$\end{tabular}}}}%
    \put(0.1295141,0.06297937){\color[rgb]{0,0,0}\makebox(0,0)[lt]{\lineheight{1.25}\smash{\begin{tabular}[t]{l}Conv\end{tabular}}}}%
    \put(0.28999731,0.09079295){\color[rgb]{0,0,0}\makebox(0,0)[lt]{\lineheight{1.25}\smash{\begin{tabular}[t]{l}$5 \times 5$\end{tabular}}}}%
    \put(0.29226891,0.06297937){\color[rgb]{0,0,0}\makebox(0,0)[lt]{\lineheight{1.25}\smash{\begin{tabular}[t]{l}Conv\end{tabular}}}}%
    \put(0.44764073,0.09079295){\color[rgb]{0,0,0}\makebox(0,0)[lt]{\lineheight{1.25}\smash{\begin{tabular}[t]{l}$1 \times 1$\end{tabular}}}}%
    \put(0.44991231,0.06297937){\color[rgb]{0,0,0}\makebox(0,0)[lt]{\lineheight{1.25}\smash{\begin{tabular}[t]{l}Conv\end{tabular}}}}%
    \put(0.59033685,0.09079295){\color[rgb]{0,0,0}\makebox(0,0)[lt]{\lineheight{1.25}\smash{\begin{tabular}[t]{l}$5 \times 5$\end{tabular}}}}%
    \put(0.59260847,0.06297937){\color[rgb]{0,0,0}\makebox(0,0)[lt]{\lineheight{1.25}\smash{\begin{tabular}[t]{l}Conv\end{tabular}}}}%
    \put(0.72423071,0.09079295){\color[rgb]{0,0,0}\makebox(0,0)[lt]{\lineheight{1.25}\smash{\begin{tabular}[t]{l}$5 \times 5$\end{tabular}}}}%
    \put(0.72650229,0.06297937){\color[rgb]{0,0,0}\makebox(0,0)[lt]{\lineheight{1.25}\smash{\begin{tabular}[t]{l}Conv\end{tabular}}}}%
    \put(0.85366412,0.09079295){\color[rgb]{0,0,0}\makebox(0,0)[lt]{\lineheight{1.25}\smash{\begin{tabular}[t]{l}$5 \times 5$\end{tabular}}}}%
    \put(0.8559357,0.06297937){\color[rgb]{0,0,0}\makebox(0,0)[lt]{\lineheight{1.25}\smash{\begin{tabular}[t]{l}Conv\end{tabular}}}}%
    \put(0.01841626,0.14835088){\color[rgb]{0.36078431,0.37254902,0.38431373}\rotatebox{90}{\makebox(0,0)[lt]{\lineheight{1.25}\smash{\begin{tabular}[t]{l}$m\times n$\end{tabular}}}}}%
    \put(0.17953604,0.33947227){\color[rgb]{0.36078431,0.37254902,0.38431373}\makebox(0,0)[lt]{\lineheight{1.25}\smash{\begin{tabular}[t]{l}$64$\end{tabular}}}}%
    \put(0.34275866,0.33947227){\color[rgb]{0.36078431,0.37254902,0.38431373}\makebox(0,0)[lt]{\lineheight{1.25}\smash{\begin{tabular}[t]{l}$64$\end{tabular}}}}%
    \put(0.50598131,0.3140539){\color[rgb]{0.36078431,0.37254902,0.38431373}\makebox(0,0)[lt]{\lineheight{1.25}\smash{\begin{tabular}[t]{l}$48$\end{tabular}}}}%
    \put(0.65580323,0.29019527){\color[rgb]{0.36078431,0.37254902,0.38431373}\makebox(0,0)[lt]{\lineheight{1.25}\smash{\begin{tabular}[t]{l}$32$\end{tabular}}}}%
    \put(0.93903272,0.26010532){\color[rgb]{0.36078431,0.37254902,0.38431373}\makebox(0,0)[lt]{\lineheight{1.25}\smash{\begin{tabular}[t]{l}$3$\end{tabular}}}}%
    \put(0.07025492,0.26010532){\color[rgb]{0.36078431,0.37254902,0.38431373}\makebox(0,0)[lt]{\lineheight{1.25}\smash{\begin{tabular}[t]{l}$3$\end{tabular}}}}%
    \put(0.79222385,0.29019527){\color[rgb]{0.36078431,0.37254902,0.38431373}\makebox(0,0)[lt]{\lineheight{1.25}\smash{\begin{tabular}[t]{l}$32$\end{tabular}}}}%
    \put(0,0){\includegraphics[width=\unitlength,page=2]{NW-Struct_public.pdf}}%
    \put(0.14224134,0.01179239){\color[rgb]{0,0,0}\makebox(0,0)[lt]{\begin{minipage}{0.22445186\unitlength}\raggedright \end{minipage}}}%
    \put(0.12485491,0.00504198){\color[rgb]{0.36078431,0.37254902,0.38431373}\makebox(0,0)[lt]{\lineheight{1.25}\smash{\begin{tabular}[t]{l}Feature Extraction\end{tabular}}}}%
    \put(0.41760308,0.00504198){\color[rgb]{0.36078431,0.37254902,0.38431373}\makebox(0,0)[lt]{\lineheight{1.25}\smash{\begin{tabular}[t]{l}Transform\end{tabular}}}}%
    \put(0.62183295,0.00504198){\color[rgb]{0.36078431,0.37254902,0.38431373}\makebox(0,0)[lt]{\lineheight{1.25}\smash{\begin{tabular}[t]{l}Image Reconstruction\end{tabular}}}}%
  \end{picture}%
\endgroup%

%% file: Images/filter_0perc_viridis.pdf_tex
\begingroup%
  \makeatletter%
  \providecommand\color[2][]{%
    \errmessage{(Inkscape) Color is used for the text in Inkscape, but the package 'color.sty' is not loaded}%
    \renewcommand\color[2][]{}%
  }%
  \providecommand\transparent[1]{%
    \errmessage{(Inkscape) Transparency is used (non-zero) for the text in Inkscape, but the package 'transparent.sty' is not loaded}%
    \renewcommand\transparent[1]{}%
  }%
  \providecommand\rotatebox[2]{#2}%
  \newcommand*\fsize{\dimexpr\f@size pt\relax}%
  \newcommand*\lineheight[1]{\fontsize{\fsize}{#1\fsize}\selectfont}%
  \ifx\svgwidth\undefined%
    \setlength{\unitlength}{648bp}%
    \ifx\svgscale\undefined%
      \relax%
    \else%
      \setlength{\unitlength}{\unitlength * \real{\svgscale}}%
    \fi%
  \else%
    \setlength{\unitlength}{\svgwidth}%
  \fi%
  \global\let\svgwidth\undefined%
  \global\let\svgscale\undefined%
  \makeatother%
  \begin{picture}(1,1)%
    \lineheight{1}%
    \setlength\tabcolsep{0pt}%
    \put(0,0){\includegraphics[width=\unitlength,page=1]{filter_0perc_viridis.pdf}}%
  \end{picture}%
\endgroup%

%% file: Images/filter_100perc_viridis.pdf_tex
\begingroup%
  \makeatletter%
  \providecommand\color[2][]{%
    \errmessage{(Inkscape) Color is used for the text in Inkscape, but the package 'color.sty' is not loaded}%
    \renewcommand\color[2][]{}%
  }%
  \providecommand\transparent[1]{%
    \errmessage{(Inkscape) Transparency is used (non-zero) for the text in Inkscape, but the package 'transparent.sty' is not loaded}%
    \renewcommand\transparent[1]{}%
  }%
  \providecommand\rotatebox[2]{#2}%
  \newcommand*\fsize{\dimexpr\f@size pt\relax}%
  \newcommand*\lineheight[1]{\fontsize{\fsize}{#1\fsize}\selectfont}%
  \ifx\svgwidth\undefined%
    \setlength{\unitlength}{648bp}%
    \ifx\svgscale\undefined%
      \relax%
    \else%
      \setlength{\unitlength}{\unitlength * \real{\svgscale}}%
    \fi%
  \else%
    \setlength{\unitlength}{\svgwidth}%
  \fi%
  \global\let\svgwidth\undefined%
  \global\let\svgscale\undefined%
  \makeatother%
  \begin{picture}(1,1)%
    \lineheight{1}%
    \setlength\tabcolsep{0pt}%
    \put(0,0){\includegraphics[width=\unitlength,page=1]{filter_100perc_viridis.pdf}}%
  \end{picture}%
\endgroup%

%% file: Images/filtereffects-6rd-4images.pdf_tex
\begingroup%
  \makeatletter%
  \providecommand\color[2][]{%
    \errmessage{(Inkscape) Color is used for the text in Inkscape, but the package 'color.sty' is not loaded}%
    \renewcommand\color[2][]{}%
  }%
  \providecommand\transparent[1]{%
    \errmessage{(Inkscape) Transparency is used (non-zero) for the text in Inkscape, but the package 'transparent.sty' is not loaded}%
    \renewcommand\transparent[1]{}%
  }%
  \providecommand\rotatebox[2]{#2}%
  \newcommand*\fsize{\dimexpr\f@size pt\relax}%
  \newcommand*\lineheight[1]{\fontsize{\fsize}{#1\fsize}\selectfont}%
  \ifx\svgwidth\undefined%
    \setlength{\unitlength}{71.99999746bp}%
    \ifx\svgscale\undefined%
      \relax%
    \else%
      \setlength{\unitlength}{\unitlength * \real{\svgscale}}%
    \fi%
  \else%
    \setlength{\unitlength}{\svgwidth}%
  \fi%
  \global\let\svgwidth\undefined%
  \global\let\svgscale\undefined%
  \makeatother%
  \begin{picture}(1,5.4)%
    \lineheight{1}%
    \setlength\tabcolsep{0pt}%
    \put(0,0){\includegraphics[width=\unitlength,page=1]{filtereffects-6rd-4images.pdf}}%
  \end{picture}%
\endgroup%

%% file: Images/filtereffects-6rdIRCNN_filters.pdf_tex
\begingroup%
  \makeatletter%
  \providecommand\color[2][]{%
    \errmessage{(Inkscape) Color is used for the text in Inkscape, but the package 'color.sty' is not loaded}%
    \renewcommand\color[2][]{}%
  }%
  \providecommand\transparent[1]{%
    \errmessage{(Inkscape) Transparency is used (non-zero) for the text in Inkscape, but the package 'transparent.sty' is not loaded}%
    \renewcommand\transparent[1]{}%
  }%
  \providecommand\rotatebox[2]{#2}%
  \newcommand*\fsize{\dimexpr\f@size pt\relax}%
  \newcommand*\lineheight[1]{\fontsize{\fsize}{#1\fsize}\selectfont}%
  \ifx\svgwidth\undefined%
    \setlength{\unitlength}{468bp}%
    \ifx\svgscale\undefined%
      \relax%
    \else%
      \setlength{\unitlength}{\unitlength * \real{\svgscale}}%
    \fi%
  \else%
    \setlength{\unitlength}{\svgwidth}%
  \fi%
  \global\let\svgwidth\undefined%
  \global\let\svgscale\undefined%
  \makeatother%
  \begin{picture}(1,0.15384615)%
    \lineheight{1}%
    \setlength\tabcolsep{0pt}%
    \put(0,0){\includegraphics[width=\unitlength,page=1]{filtereffects-6rdIRCNN_filters.pdf}}%
  \end{picture}%
\endgroup%

%% file: Images/filtereffects-6rdIRCNN_images.pdf_tex
\begingroup%
  \makeatletter%
  \providecommand\color[2][]{%
    \errmessage{(Inkscape) Color is used for the text in Inkscape, but the package 'color.sty' is not loaded}%
    \renewcommand\color[2][]{}%
  }%
  \providecommand\transparent[1]{%
    \errmessage{(Inkscape) Transparency is used (non-zero) for the text in Inkscape, but the package 'transparent.sty' is not loaded}%
    \renewcommand\transparent[1]{}%
  }%
  \providecommand\rotatebox[2]{#2}%
  \newcommand*\fsize{\dimexpr\f@size pt\relax}%
  \newcommand*\lineheight[1]{\fontsize{\fsize}{#1\fsize}\selectfont}%
  \ifx\svgwidth\undefined%
    \setlength{\unitlength}{467.99998347bp}%
    \ifx\svgscale\undefined%
      \relax%
    \else%
      \setlength{\unitlength}{\unitlength * \real{\svgscale}}%
    \fi%
  \else%
    \setlength{\unitlength}{\svgwidth}%
  \fi%
  \global\let\svgwidth\undefined%
  \global\let\svgscale\undefined%
  \makeatother%
  \begin{picture}(1,0.83076923)%
    \lineheight{1}%
    \setlength\tabcolsep{0pt}%
    \put(0,0){\includegraphics[width=\unitlength,page=1]{filtereffects-6rdIRCNN_images.pdf}}%
  \end{picture}%
\endgroup%

%% file: Images/filtereffects-6rd-4GBCNN_filters.pdf_tex
\begingroup%
  \makeatletter%
  \providecommand\color[2][]{%
    \errmessage{(Inkscape) Color is used for the text in Inkscape, but the package 'color.sty' is not loaded}%
    \renewcommand\color[2][]{}%
  }%
  \providecommand\transparent[1]{%
    \errmessage{(Inkscape) Transparency is used (non-zero) for the text in Inkscape, but the package 'transparent.sty' is not loaded}%
    \renewcommand\transparent[1]{}%
  }%
  \providecommand\rotatebox[2]{#2}%
  \newcommand*\fsize{\dimexpr\f@size pt\relax}%
  \newcommand*\lineheight[1]{\fontsize{\fsize}{#1\fsize}\selectfont}%
  \ifx\svgwidth\undefined%
    \setlength{\unitlength}{468bp}%
    \ifx\svgscale\undefined%
      \relax%
    \else%
      \setlength{\unitlength}{\unitlength * \real{\svgscale}}%
    \fi%
  \else%
    \setlength{\unitlength}{\svgwidth}%
  \fi%
  \global\let\svgwidth\undefined%
  \global\let\svgscale\undefined%
  \makeatother%
  \begin{picture}(1,0.15384615)%
    \lineheight{1}%
    \setlength\tabcolsep{0pt}%
    \put(0,0){\includegraphics[width=\unitlength,page=1]{filtereffects-6rd-4GBCNN_filters.pdf}}%
  \end{picture}%
\endgroup%

%% file: Images/filtereffects-6rd-4GBCNN_images.pdf_tex
\begingroup%
  \makeatletter%
  \providecommand\color[2][]{%
    \errmessage{(Inkscape) Color is used for the text in Inkscape, but the package 'color.sty' is not loaded}%
    \renewcommand\color[2][]{}%
  }%
  \providecommand\transparent[1]{%
    \errmessage{(Inkscape) Transparency is used (non-zero) for the text in Inkscape, but the package 'transparent.sty' is not loaded}%
    \renewcommand\transparent[1]{}%
  }%
  \providecommand\rotatebox[2]{#2}%
  \newcommand*\fsize{\dimexpr\f@size pt\relax}%
  \newcommand*\lineheight[1]{\fontsize{\fsize}{#1\fsize}\selectfont}%
  \ifx\svgwidth\undefined%
    \setlength{\unitlength}{467.99998347bp}%
    \ifx\svgscale\undefined%
      \relax%
    \else%
      \setlength{\unitlength}{\unitlength * \real{\svgscale}}%
    \fi%
  \else%
    \setlength{\unitlength}{\svgwidth}%
  \fi%
  \global\let\svgwidth\undefined%
  \global\let\svgscale\undefined%
  \makeatother%
  \begin{picture}(1,0.83076923)%
    \lineheight{1}%
    \setlength\tabcolsep{0pt}%
    \put(0,0){\includegraphics[width=\unitlength,page=1]{filtereffects-6rd-4GBCNN_images.pdf}}%
  \end{picture}%
\endgroup%

%% file: Images/setupcomparison_reduced_5000_peak_signal_noise_ratio_validation.pdf_tex
\begingroup%
  \makeatletter%
  \providecommand\color[2][]{%
    \errmessage{(Inkscape) Color is used for the text in Inkscape, but the package 'color.sty' is not loaded}%
    \renewcommand\color[2][]{}%
  }%
  \providecommand\transparent[1]{%
    \errmessage{(Inkscape) Transparency is used (non-zero) for the text in Inkscape, but the package 'transparent.sty' is not loaded}%
    \renewcommand\transparent[1]{}%
  }%
  \providecommand\rotatebox[2]{#2}%
  \newcommand*\fsize{\dimexpr\f@size pt\relax}%
  \newcommand*\lineheight[1]{\fontsize{\fsize}{#1\fsize}\selectfont}%
  \ifx\svgwidth\undefined%
    \setlength{\unitlength}{1028.20873273bp}%
    \ifx\svgscale\undefined%
      \relax%
    \else%
      \setlength{\unitlength}{\unitlength * \real{\svgscale}}%
    \fi%
  \else%
    \setlength{\unitlength}{\svgwidth}%
  \fi%
  \global\let\svgwidth\undefined%
  \global\let\svgscale\undefined%
  \makeatother%
  \begin{picture}(1,0.38263971)%
    \lineheight{1}%
    \setlength\tabcolsep{0pt}%
    \put(0,0){\includegraphics[width=\unitlength,page=1]{setupcomparison_reduced_5000_peak_signal_noise_ratio_validation.pdf}}%
  \end{picture}%
\endgroup%

%% file: Images/filterdevel_test_test20-088_numbers.pdf_tex
\begingroup%
  \makeatletter%
  \providecommand\color[2][]{%
    \errmessage{(Inkscape) Color is used for the text in Inkscape, but the package 'color.sty' is not loaded}%
    \renewcommand\color[2][]{}%
  }%
  \providecommand\transparent[1]{%
    \errmessage{(Inkscape) Transparency is used (non-zero) for the text in Inkscape, but the package 'transparent.sty' is not loaded}%
    \renewcommand\transparent[1]{}%
  }%
  \providecommand\rotatebox[2]{#2}%
  \newcommand*\fsize{\dimexpr\f@size pt\relax}%
  \newcommand*\lineheight[1]{\fontsize{\fsize}{#1\fsize}\selectfont}%
  \ifx\svgwidth\undefined%
    \setlength{\unitlength}{547.2000246bp}%
    \ifx\svgscale\undefined%
      \relax%
    \else%
      \setlength{\unitlength}{\unitlength * \real{\svgscale}}%
    \fi%
  \else%
    \setlength{\unitlength}{\svgwidth}%
  \fi%
  \global\let\svgwidth\undefined%
  \global\let\svgscale\undefined%
  \makeatother%
  \begin{picture}(1,0.03947368)%
    \lineheight{1}%
    \setlength\tabcolsep{0pt}%
    \put(0,0){\includegraphics[width=\unitlength,page=1]{filterdevel_test_test20-088_numbers.pdf}}%
  \end{picture}%
\endgroup%

%% file: Images/filterdevel_test_test20-088_8.pdf_tex
\begingroup%
  \makeatletter%
  \providecommand\color[2][]{%
    \errmessage{(Inkscape) Color is used for the text in Inkscape, but the package 'color.sty' is not loaded}%
    \renewcommand\color[2][]{}%
  }%
  \providecommand\transparent[1]{%
    \errmessage{(Inkscape) Transparency is used (non-zero) for the text in Inkscape, but the package 'transparent.sty' is not loaded}%
    \renewcommand\transparent[1]{}%
  }%
  \providecommand\rotatebox[2]{#2}%
  \newcommand*\fsize{\dimexpr\f@size pt\relax}%
  \newcommand*\lineheight[1]{\fontsize{\fsize}{#1\fsize}\selectfont}%
  \ifx\svgwidth\undefined%
    \setlength{\unitlength}{547.20001351bp}%
    \ifx\svgscale\undefined%
      \relax%
    \else%
      \setlength{\unitlength}{\unitlength * \real{\svgscale}}%
    \fi%
  \else%
    \setlength{\unitlength}{\svgwidth}%
  \fi%
  \global\let\svgwidth\undefined%
  \global\let\svgscale\undefined%
  \makeatother%
  \begin{picture}(1,0.27631578)%
    \lineheight{1}%
    \setlength\tabcolsep{0pt}%
    \put(0,0){\includegraphics[width=\unitlength,page=1]{filterdevel_test_test20-088_8.pdf}}%
  \end{picture}%
\endgroup%

%% file: Images/filterdevel_test_test20-088_11.pdf_tex
\begingroup%
  \makeatletter%
  \providecommand\color[2][]{%
    \errmessage{(Inkscape) Color is used for the text in Inkscape, but the package 'color.sty' is not loaded}%
    \renewcommand\color[2][]{}%
  }%
  \providecommand\transparent[1]{%
    \errmessage{(Inkscape) Transparency is used (non-zero) for the text in Inkscape, but the package 'transparent.sty' is not loaded}%
    \renewcommand\transparent[1]{}%
  }%
  \providecommand\rotatebox[2]{#2}%
  \newcommand*\fsize{\dimexpr\f@size pt\relax}%
  \newcommand*\lineheight[1]{\fontsize{\fsize}{#1\fsize}\selectfont}%
  \ifx\svgwidth\undefined%
    \setlength{\unitlength}{547.20001351bp}%
    \ifx\svgscale\undefined%
      \relax%
    \else%
      \setlength{\unitlength}{\unitlength * \real{\svgscale}}%
    \fi%
  \else%
    \setlength{\unitlength}{\svgwidth}%
  \fi%
  \global\let\svgwidth\undefined%
  \global\let\svgscale\undefined%
  \makeatother%
  \begin{picture}(1,0.27631578)%
    \lineheight{1}%
    \setlength\tabcolsep{0pt}%
    \put(0,0){\includegraphics[width=\unitlength,page=1]{filterdevel_test_test20-088_11.pdf}}%
  \end{picture}%
\endgroup%

%% file: Images/filterdevel_test_test20-088_21.pdf_tex
\begingroup%
  \makeatletter%
  \providecommand\color[2][]{%
    \errmessage{(Inkscape) Color is used for the text in Inkscape, but the package 'color.sty' is not loaded}%
    \renewcommand\color[2][]{}%
  }%
  \providecommand\transparent[1]{%
    \errmessage{(Inkscape) Transparency is used (non-zero) for the text in Inkscape, but the package 'transparent.sty' is not loaded}%
    \renewcommand\transparent[1]{}%
  }%
  \providecommand\rotatebox[2]{#2}%
  \newcommand*\fsize{\dimexpr\f@size pt\relax}%
  \newcommand*\lineheight[1]{\fontsize{\fsize}{#1\fsize}\selectfont}%
  \ifx\svgwidth\undefined%
    \setlength{\unitlength}{547.20001351bp}%
    \ifx\svgscale\undefined%
      \relax%
    \else%
      \setlength{\unitlength}{\unitlength * \real{\svgscale}}%
    \fi%
  \else%
    \setlength{\unitlength}{\svgwidth}%
  \fi%
  \global\let\svgwidth\undefined%
  \global\let\svgscale\undefined%
  \makeatother%
  \begin{picture}(1,0.27631578)%
    \lineheight{1}%
    \setlength\tabcolsep{0pt}%
    \put(0,0){\includegraphics[width=\unitlength,page=1]{filterdevel_test_test20-088_21.pdf}}%
  \end{picture}%
\endgroup%

%% file: Images/filterdevel_test_test20-088_40.pdf_tex
\begingroup%
  \makeatletter%
  \providecommand\color[2][]{%
    \errmessage{(Inkscape) Color is used for the text in Inkscape, but the package 'color.sty' is not loaded}%
    \renewcommand\color[2][]{}%
  }%
  \providecommand\transparent[1]{%
    \errmessage{(Inkscape) Transparency is used (non-zero) for the text in Inkscape, but the package 'transparent.sty' is not loaded}%
    \renewcommand\transparent[1]{}%
  }%
  \providecommand\rotatebox[2]{#2}%
  \newcommand*\fsize{\dimexpr\f@size pt\relax}%
  \newcommand*\lineheight[1]{\fontsize{\fsize}{#1\fsize}\selectfont}%
  \ifx\svgwidth\undefined%
    \setlength{\unitlength}{547.20001351bp}%
    \ifx\svgscale\undefined%
      \relax%
    \else%
      \setlength{\unitlength}{\unitlength * \real{\svgscale}}%
    \fi%
  \else%
    \setlength{\unitlength}{\svgwidth}%
  \fi%
  \global\let\svgwidth\undefined%
  \global\let\svgscale\undefined%
  \makeatother%
  \begin{picture}(1,0.27631578)%
    \lineheight{1}%
    \setlength\tabcolsep{0pt}%
    \put(0,0){\includegraphics[width=\unitlength,page=1]{filterdevel_test_test20-088_40.pdf}}%
  \end{picture}%
\endgroup%

%% file: Images/filterdevel_test_test20-088_57.pdf_tex
\begingroup%
  \makeatletter%
  \providecommand\color[2][]{%
    \errmessage{(Inkscape) Color is used for the text in Inkscape, but the package 'color.sty' is not loaded}%
    \renewcommand\color[2][]{}%
  }%
  \providecommand\transparent[1]{%
    \errmessage{(Inkscape) Transparency is used (non-zero) for the text in Inkscape, but the package 'transparent.sty' is not loaded}%
    \renewcommand\transparent[1]{}%
  }%
  \providecommand\rotatebox[2]{#2}%
  \newcommand*\fsize{\dimexpr\f@size pt\relax}%
  \newcommand*\lineheight[1]{\fontsize{\fsize}{#1\fsize}\selectfont}%
  \ifx\svgwidth\undefined%
    \setlength{\unitlength}{547.20001351bp}%
    \ifx\svgscale\undefined%
      \relax%
    \else%
      \setlength{\unitlength}{\unitlength * \real{\svgscale}}%
    \fi%
  \else%
    \setlength{\unitlength}{\svgwidth}%
  \fi%
  \global\let\svgwidth\undefined%
  \global\let\svgscale\undefined%
  \makeatother%
  \begin{picture}(1,0.27631578)%
    \lineheight{1}%
    \setlength\tabcolsep{0pt}%
    \put(0,0){\includegraphics[width=\unitlength,page=1]{filterdevel_test_test20-088_57.pdf}}%
  \end{picture}%
\endgroup%

%% file: Images/filterdevel_test_test20-088_58.pdf_tex
\begingroup%
  \makeatletter%
  \providecommand\color[2][]{%
    \errmessage{(Inkscape) Color is used for the text in Inkscape, but the package 'color.sty' is not loaded}%
    \renewcommand\color[2][]{}%
  }%
  \providecommand\transparent[1]{%
    \errmessage{(Inkscape) Transparency is used (non-zero) for the text in Inkscape, but the package 'transparent.sty' is not loaded}%
    \renewcommand\transparent[1]{}%
  }%
  \providecommand\rotatebox[2]{#2}%
  \newcommand*\fsize{\dimexpr\f@size pt\relax}%
  \newcommand*\lineheight[1]{\fontsize{\fsize}{#1\fsize}\selectfont}%
  \ifx\svgwidth\undefined%
    \setlength{\unitlength}{547.20001351bp}%
    \ifx\svgscale\undefined%
      \relax%
    \else%
      \setlength{\unitlength}{\unitlength * \real{\svgscale}}%
    \fi%
  \else%
    \setlength{\unitlength}{\svgwidth}%
  \fi%
  \global\let\svgwidth\undefined%
  \global\let\svgscale\undefined%
  \makeatother%
  \begin{picture}(1,0.27631578)%
    \lineheight{1}%
    \setlength\tabcolsep{0pt}%
    \put(0,0){\includegraphics[width=\unitlength,page=1]{filterdevel_test_test20-088_58.pdf}}%
  \end{picture}%
\endgroup%

%% file: Images/trainingsample-evolution_00-25_reduced-two_peak_signal_noise_ratio_validation_xlog.pdf_tex
\begingroup%
  \makeatletter%
  \providecommand\color[2][]{%
    \errmessage{(Inkscape) Color is used for the text in Inkscape, but the package 'color.sty' is not loaded}%
    \renewcommand\color[2][]{}%
  }%
  \providecommand\transparent[1]{%
    \errmessage{(Inkscape) Transparency is used (non-zero) for the text in Inkscape, but the package 'transparent.sty' is not loaded}%
    \renewcommand\transparent[1]{}%
  }%
  \providecommand\rotatebox[2]{#2}%
  \newcommand*\fsize{\dimexpr\f@size pt\relax}%
  \newcommand*\lineheight[1]{\fontsize{\fsize}{#1\fsize}\selectfont}%
  \ifx\svgwidth\undefined%
    \setlength{\unitlength}{1028.20873979bp}%
    \ifx\svgscale\undefined%
      \relax%
    \else%
      \setlength{\unitlength}{\unitlength * \real{\svgscale}}%
    \fi%
  \else%
    \setlength{\unitlength}{\svgwidth}%
  \fi%
  \global\let\svgwidth\undefined%
  \global\let\svgscale\undefined%
  \makeatother%
  \begin{picture}(1,0.38715849)%
    \lineheight{1}%
    \setlength\tabcolsep{0pt}%
    \put(0,0){\includegraphics[width=\unitlength,page=1]{trainingsample-evolution_00-25_reduced-two_peak_signal_noise_ratio_validation_xlog.pdf}}%
  \end{picture}%
\endgroup%